\documentclass[sigconf]{acmart}

\usepackage{graphicx} 
\usepackage{booktabs} 
\usepackage{multirow} 
\usepackage{amsmath}  
\usepackage{subcaption} 
\usepackage{algorithm}    
\usepackage{algorithmic}
\usepackage{color}
\usepackage{xcolor}
\usepackage{amsmath}
\newtheorem{theorem}{Theorem}
\newtheorem{lemma}{Lemma}
\usepackage[encapsulated]{CJK}
\usepackage{makecell}
\newcommand{\chinese}[1]{\begin{CJK}{UTF8}{gbsn} #1 \end{CJK}}
\definecolor{cyan}{RGB}{154,201,219}
\definecolor{flesh}{RGB}{248,172,140}
\usepackage{tcolorbox}
\usepackage{verbatim}
\usepackage{listings}
\usepackage{courier}
\tcbuselibrary{listings, breakable}

\lstdefinelanguage{plainjson}{
  basicstyle=\ttfamily\fontsize{5}{6},
  breaklines=true,
  showstringspaces=false,
  escapeinside={(*@}{@*)} 
}

\AtBeginDocument{%
  }

\copyrightyear{2025}
\acmYear{2025}
\setcopyright{acmlicensed}
\acmConference[CIKM '25]{Proceedings of the 34th ACM International Conference on Information and Knowledge Management}{November 10--14, 2025}{Seoul, Republic of Korea}
\acmBooktitle{Proceedings of the 34th ACM International Conference on Information and Knowledge Management (CIKM '25), November 10--14, 2025,
Seoul, Republic of Korea}
\acmDOI{10.1145/3746252.3761389}
\acmISBN{979-8-4007-2040-6/2025/11}

\begin{document}

\title{Querier-Aware LLM: Generating Personalized Responses to the Same Query from Different Queriers}

\author{Hang Zeng}
\orcid{0009-0009-6841-9872}
\affiliation{%
  \institution{Shanghai Jiao Tong University}
  \city{Shanghai}
  \country{China}}
\email{nidhogg@sjtu.edu.cn}

\author{Chaoyue Niu}
\authornote{Chaoyue Niu is the corresponding author.}
\orcid{0000-0002-1650-4233}
\affiliation{%
  \institution{Shanghai Jiao Tong University}
  \city{Shanghai}
  \country{China}
}
\email{rvince@sjtu.edu.cn}

\author{Fan Wu}
\orcid{0000-0003-0965-9058}
\affiliation{%
  \institution{Shanghai Jiao Tong University}
  \city{Shanghai}
  \country{China}
}
\email{wu-fan@sjtu.edu.cn}

\author{Chengfei Lyu}
\orcid{0009-0004-6618-521X}
\affiliation{%
  \institution{Alibaba Group}
  \city{Hangzhou, Zhejiang}
  \country{China}
}
\email{chengfei.lcf@alibaba-inc.com}

\author{Guihai Chen}
\orcid{0000-0002-6934-1685}
\affiliation{%
  \institution{Shanghai Jiao Tong University}
  \city{Shanghai}
  \country{China}
}
\email{gchen@cs.sjtu.edu.cn}

\renewcommand{\shortauthors}{Zeng et al.}

\begin{abstract}
Existing work on large language model (LLM) personalization assigned different responding roles to LLMs, but overlooked the diversity of queriers. In this work, we propose a new form of querier-aware LLM personalization, generating different responses even for the same query from different queriers. We design a dual-tower model architecture with a cross-querier general encoder and a querier-specific encoder. We further apply contrastive learning with multi-view augmentation, pulling close the dialogue representations of the same querier, while pulling apart those of different queriers. To mitigate the impact of query diversity on querier-contrastive learning, we cluster the dialogues based on query similarity and restrict the scope of contrastive learning within each cluster. To address the lack of datasets designed for querier-aware personalization, we also build a multi-querier dataset from English and Chinese scripts, as well as WeChat records, called MQDialog, containing 173 queriers and 12 responders. Extensive evaluations demonstrate that our design significantly improves the quality of personalized response generation, achieving relative improvement of 8.4\% to 48.7\% in ROUGE-L scores and winning rates ranging from 54\% to 82\% compared with various baseline methods\footnote{The code is public at \href{https://github.com/Nidryen-zh/QuerierAwareResponder}{https://github.com/Nidryen-zh/QuerierAwareResponder} and the dataset can be found at \href{https://huggingface.co/datasets/Nidhogg-zh/Multi-Querier_Dialogue}{https://huggingface.co/datasets/Nidhogg-zh/Multi-Querier$\_$Dialogue}.}. 
\end{abstract}





\maketitle

\section{Introduction}

\begin{figure}[!t]
  \centering
  \vspace{0.8em}
    \begin{subfigure}{0.9\linewidth}
         \centering
         \includegraphics[width=\textwidth]{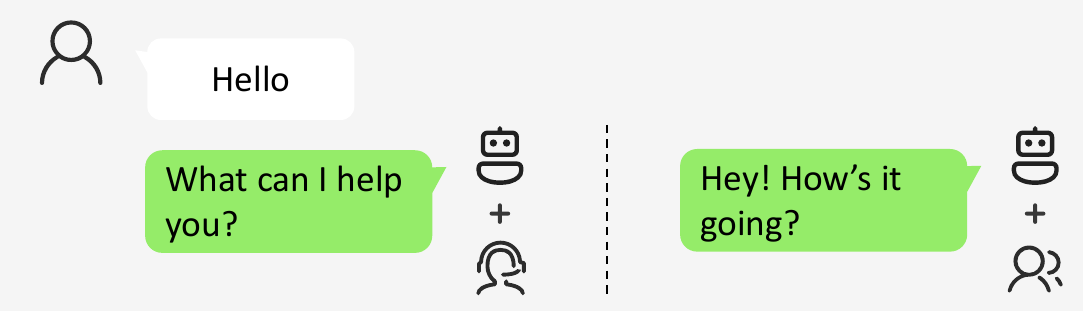}
         \caption{\centering Responder in Different Roles}
         \label{fig:traindional}
    \end{subfigure}
    \\[1ex]
    \begin{subfigure}{0.9\linewidth}
         \centering
         \includegraphics[width=\textwidth]{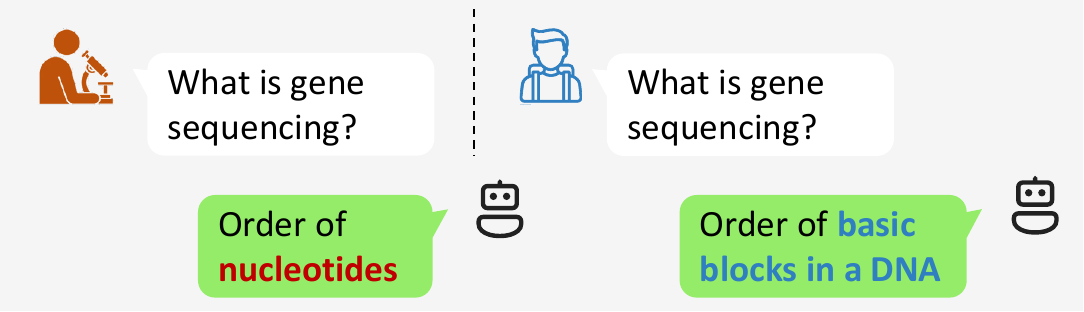}
         \caption{\centering Querier-Aware Responder}
         \label{fig:qsa}
    \end{subfigure}
    \caption{LLM personalization from responder-side diversity in existing work and from querier-side diversity in this work.}
  \label{fig:personalized_type}
  \vspace{-0.8em}
\end{figure}

Generative LLM has revolutionized and streamlined how users interact with artificial intelligence. When users pose queries, LLM understands, interprets, and generates responses with unprecedented depth and contextual nuance. To offer diverse interaction styles, existing work assigned roles with domain-specific knowledge to LLM \cite{dialogueagent, characterllm, rolellm}. As shown in Figure \ref{fig:traindional}, when a user greets with ``Hello'', LLM in the role of a helpful assistant might reply ``What can I help you?'', while LLM with a different role of a good friend could respond more casually with ``Hey! How’s it going?''.

Parallel to existing works on responder-side diversity, we study a new form of querier-aware personalization, where the LLM adapts to a diverse range of users who pose queries. The desired effect is that even for the same query from different users, LLM will generate customized responses that are tailored to each querier's unique personality, their mutual relationship with the responder, as well as the conversational context. In the intricate tapestry of human communication, our conversational stance, linguistic choices, and emotional inflections are profoundly shaped by the specific person with whom we are engaging. The concept of querier-aware responding can also be traced back to the sentiment expressed by Baojia Li in the novel The Appearance of Officialdom (Qing Dynasty), which states: ``When you meet a person, speak to them as a person; when you meet a ghost, speak to it as a ghost; when you’re among officials, speak the language of officialdom; and when you’re among businessmen, speak the language of the business world.''

Querier-aware LLM personalization is important in practical applications. Consider an intelligent assistant for answering the query about genetic sequencing, as shown in Figure \ref{fig:qsa}. To a bioinformatician, the response requires sophisticated technical vocabulary, advanced methodological details, and domain-specific references. In contrast, when a high school student poses the same query, it requires a carefully crafted narrative that simplifies complex concepts, employs accessible analogies, and progressively builds foundational understanding. By adapting to the knowledge levels of queriers, LLMs can enhance information comprehension and absorption. 
Moreover, consider the reporting of an issue in LLM-based automatic operations and maintenance. To engineers, the response should emphasize technical details and necessary actions, using a solution-oriented and jargon-rich tone. To external customers, the response should prioritize reassurance and transparency regarding service quality, adopting an empathetic and customer-centric tone. By fostering querier-awareness, LLM can more effectively address specific information needs.

To achieve querier-aware responses, trivially prompting LLM with the querier's profile or a few dialogue history does not fully leverage the potential of interaction history and cannot deeply represent users in LLM for personalization. Additionally, tuning a querier-specific LLM for each user over their conversation history is costly, unsustainable for large user bases, and fails to exploit all users' interactions during training for the current user.

In this work, we consider querier-aware LLM personalization that incorporates each querier's personality and their relationships with a responder by exploiting all the dialogues between queriers and the responder. The proposed model architecture comprises a cross-querier general encoder initialized with a pretrained LLM and a querier-specific encoder characterized by a low-rank intrinsic property. To differentiate between the representations of various queriers, we introduce a querier-contrastive loss to guide model training. In particular, we group dialogues containing similar queries from different queriers into clusters, and then construct positive and negative pairs within each cluster for contrastive learning. We further propose a multi-view augmentation strategy to enhance the effectiveness and efficiency of contrastive learning. Considering that existing datasets cannot meet querier-aware evaluation requirements, we build a multi-querier dialogue dataset (MQDialog) by extracting and cleaning the dialogues between different queriers and a certain responder. The source data includes English and Chinese scripts, as well as real-world WeChat records. 

We summarize the key contributions as follows:
(1) We study a new form of LLM personalization with querier-aware response generation and construct a MQDialog dataset for benchmark testing purposes.
(2) We design a dual-tower model architecture, a querier-aware contrastive loss function with multi-view augmentation, and a query similarity-based dialogue clustering strategy for restricting contrastive learning scope.
(3) Evaluation results demonstrate that our design achieves an average relative improvement of at least 8.6\% in BLEU and 8.4\% in ROUGE-L scores, along with a 65.8\% average winning rate judged by GPT-4, compared to various baselines.

\section{Related Work}
Most existing work focuses on responder-side personalization, where different personalities are assigned to the LLM \cite{personalizeddialogue}, such as role-playing \cite{frompersona, roleplay1, roleplay2}. Some works focus on maintaining role consistency in generated responses \cite{hamlets, personalizedknowledge, roleeval}, by finetuning LLM on carefully crafted role profiles and instructions \cite{characterllm, exploitingpersona}, or by using retrieval augmentation methods \cite{learningretrieval, hallucination}. Other works emphasize personalized tone and linguistic features of roles, using supervised finetuning or reinforcement learning on dialogue datasets \cite{characterglm}, integrating role profiles \cite{harrypotter}, or incorporating background stories into the input \cite{chatharuhi}. From a broader perspective, personalized text generation \cite{personalizedresponse} can also be viewed as a specific instance of role-play. Existing work on personalized review generation \cite{livechat, trainingmillions} enables LLMs to emulate as reviewers through embedding responder-specific personality. However, these studies focused on the personality of the assigned role, without modeling personalized responses tailored to the diverse characteristics of queriers.

Another line of work is user information-based generation, typically, in chatbots \cite{onechatbot, surveyperson} and recommendation engines \cite{surveyrec, personrec}, including dialogue history-based LLM \cite{responsegeneration, helloagain}, profile-based LLM \cite{onechatbot, prompt-based, prompt-based3}, and incorporating user embedding into models \cite{user-emb1, user-emb2}. However, these works aggregate all users' information together for training \cite{oppu}, learning response patterns for specific user information, such as dialogue content, and treat users independently \cite{lamp}. 
They generate specific responses based on different user information, rather than explicitly modeling the user's unique personality. 

In the field of representation learning, representation alignment ensures consistency across different perspectives of the data, while representation differentiation focuses on extracting the unique traits of each sample. One prominent method for achieving representation alignment is multi-view learning \cite{multiviewsurvey}. Existing work leverages the correlation between different views \cite{relationsbetween, deepcca, deepmultiview} and the distance \cite{devise} between samples to integrate information across diverse views. View reconstruction \cite{partialmultiview} and view augmentation \cite{multiviewinformation} are also helpful in enhancing alignment. On the other hand, contrastive learning is a key method for representation differentiation, focusing on maximizing the similarity between positive pairs and minimizing it for negative pairs \cite{distancemetric, cpc, dim}. Some work extends contrastive learning to multiple views \cite{maximizingmi, cmc}, pulling features of different views from the same sample closer while pushing apart features of views from different samples. 

\section{Problem Formulation}\label{sec:problem}

\subsection{Querier-Aware Response Generation}\label{sec:fprp}

We let $\mathbb{Q} = \{Q^1, Q^2, \ldots, Q^m\}$ denote $m$ different queriers and refer to LLM as the responder $R$. For each querier $Q^i$, we let $\mathcal{D}^i =\{d^{i}_1, d^{i}_2, \ldots, d^{i}_{n_i}\}$ denote a collection of $n_i$ dialogues with $R$. Each dialogue $d^{i}_k=(x^{i}_k, y^{i}_k)$ consists of $x^{i}_k$, the context history, typically including multiple rounds of previous interactions and the latest query, and $y^{i}_k$, the corresponding ground-truth response. Querier-aware response generation desires that for any two different queriers $Q^i \neq Q^j$, given the same query context history $x^{i}_k = x^{j}_l$, the distributions of generated responses differ, formalized as 
\begin{equation}\label{eq:objective}
    P\left(y^{i}_k|x^{i}_k; Q^i; \Theta\right) \neq P\left(y^{j}_l|x^{j}_l; Q^j; \Theta\right),
\end{equation}
where $\Theta$ denotes the parameters of LLM. The optimization objective of querier-aware personalized LLM is still to maximize the overall likelihood across all queriers and dialogues, formalized as
\begin{equation}
\Theta = \mathop{\arg\max}\limits_{\Theta} \prod_{i=1}^{m} \prod_{k=1}^{n_{i}} P\left({y}^{i}_k|{x}^{i}_k; Q^i;\Theta\right).    
\end{equation}

\subsection{Trivial Methods and Our Consideration}\label{sec:trivial_tune}

One straightforward way to enable LLM to distinguish between different queriers $\mathbb{Q}$ is to incorporate querier-specific information, such as user profiles, into the prompt. Additionally, dialogue histories associated with a specific querier can be utilized as supplementary input for few-shot prompting or in-context learning. However, the key problem is that they do not fully exploit all the dialogues to deeply and comprehensively encode each individual querier's features into the LLM. Moreover, it is costly to build fine-grained profiles for a large number of users in practice. Meanwhile, static user profiles and fixed prompting examples cannot adapt effectively to varying conversational contexts in the real world. 

To integrate querier-specific response capability into LLM, a trivial method is to train a personalized LLM for each individual querier $Q^i$, based on the dialogue history between $Q^i$ and $R$. When serving $Q^i$, the corresponding personalized LLM is invoked to generate responses. The key problem is that the rich dialogues with other queriers $\mathbb{Q} \setminus \{Q_i\}$ are not exploited to capture the general personality of $R$ which is inherently embedded in the dialogues with all queriers $\mathbb{Q}$, and the potential benefit of contrastive learning against other queriers is also neglected. Furthermore, for a large user base, training personalized models and maintaining them for real-time serving incurs unaffordable offline and online overhead.

For the above concerns, we consider the new problem of how to effectively and efficiently exploit dialogues from all the queriers to integrate the querier-aware response generation ability into a unified model, intuitively, a personalized one-for-all LLM.

\section{Design}\label{sec:design}

The overall insight of our key design is to divide the personality in dialogue into two components: a general personality that represents the responder's cross-querier characteristics, and a querier-specific personality that captures the unique characteristics of individual queriers and their relationships with the responder. We introduce the design from sample construction, model architecture, and loss function. In particular, we first propose to cluster dialogues based on the semantic similarity of queries, mitigating the impact of query diversity on contrastive learning across different queriers. We then design a dual-tower architecture with a general encoder and a specific encoder. We finally perform contrastive learning with multi-view augmentation in each cluster, aligning representations for the same querier while differentiating those of different queriers.

\subsection{Query Similarity Based Dialogue Clustering}

An ideal way for sample construction is to let different users ask the same query and collect corresponding responses for querier-contrastive learning while minimizing the impact of query diversity. However, the real-world dataset with such samples does not exist in practice. To deal with this problem, we relax the strict requirement of the same query $x^{i}_k = x^{j}_l$ from $Q^i$ and $Q^j$ in equation \ref{eq:objective}, allowing instead for similar queries $x^{i}_k \approx x^{j}_l$. Thus, given an existing dialogue dataset, we cluster similar queries rather than requiring exact matches and then perform personalized LLM learning within these clusters. In particular, for each dialogue $d_k^{i}$ between $Q^i$ and $R$, the context history $x_k^{i}$ contains tokens not only from $Q^i$ but also from $R$. We filter out the tokens from $R$ and keep only the tokens from $Q^i$, thus forming a querier-oriented context $q_k^{i}$. By inputting $q_k^{i}$ into a text embedding model (e.g., OpenAI's text-embedding-ada-002 in this work), we can obtain its semantic vector. We further apply k-means clustering to group similar queries using the cosine distance between semantic vectors as the similarity measure. By clustering based on query similarity, we obtain samples from different queriers but with similar styles, enabling models to distinguish between dialogues from different queriers and facilitating querier-specific representation learning. This approach also prevents the issue of naively forcing all dialogues even with different queries from one querier to have similar representations, which is essential for effectively learning to capture semantic information.

\begin{figure}[!t]
  \centering
   \includegraphics[width=0.95\columnwidth]{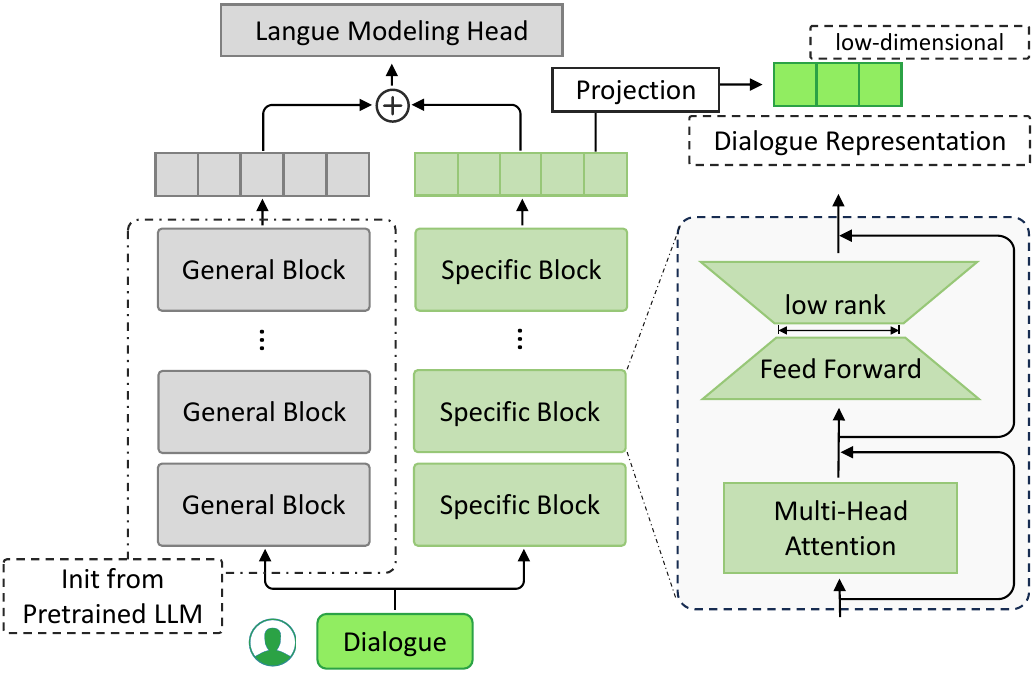}
    \caption{Dual-tower architecture. The feedfroward layer of the specific block is composed of two low-rank matrices.} \label{fig:model}
    \vspace{-0.8em}
\end{figure}

\subsection{Dual-Tower Model Architecture}\label{sec:approach}

To disentangle cross-querier general representation from querier-specific representation, we propose a dual-tower model architecture, as illustrated in Figure \ref{fig:model}. The first ``tower'' is the general encoder with multiple transformer blocks, denoted as $G_{g}$, which can be initialized with a pretrained LLM. The second ``tower'' is the specific encoder with multiple adapted transformer blocks, denoted as $G_{s}$. A key adaptation lies in the dense feedforward layer after multi-head attention, which is decomposed into two low-rank matrices. The design insight is that querier-specific representations are intrinsically sparser compared to cross-querier general representations, making low-rank matrices sufficient. In addition to effectiveness, the low-rank design also makes the learning process parameter-efficient. 
Both the general encoder and the specific encoder are shared across all queriers. Therefore, with the number of queriers growing, the model structure remains unchanged, and the training and inference cost will not increase.
Two representations after the general encoder and the specific encoder are further fused using element-wise addition and then passed through the language model header to predict the next token.

The training of $G_{g}$ is guided by the language modeling loss \cite{elmo}, while the training of $G_{s}$ is guided by both the language modeling loss and the querier-contrastive loss with multi-view augmentation defined in equation \ref{eq:loss:qc:mv}. Additionally, contrastive learning is conducted among each cluster of dialogues containing similar queries from different queriers.   

\subsection{Querier-Contrastive Loss}

\begin{figure}[!t]
  \centering
    \includegraphics[width=0.95\columnwidth]{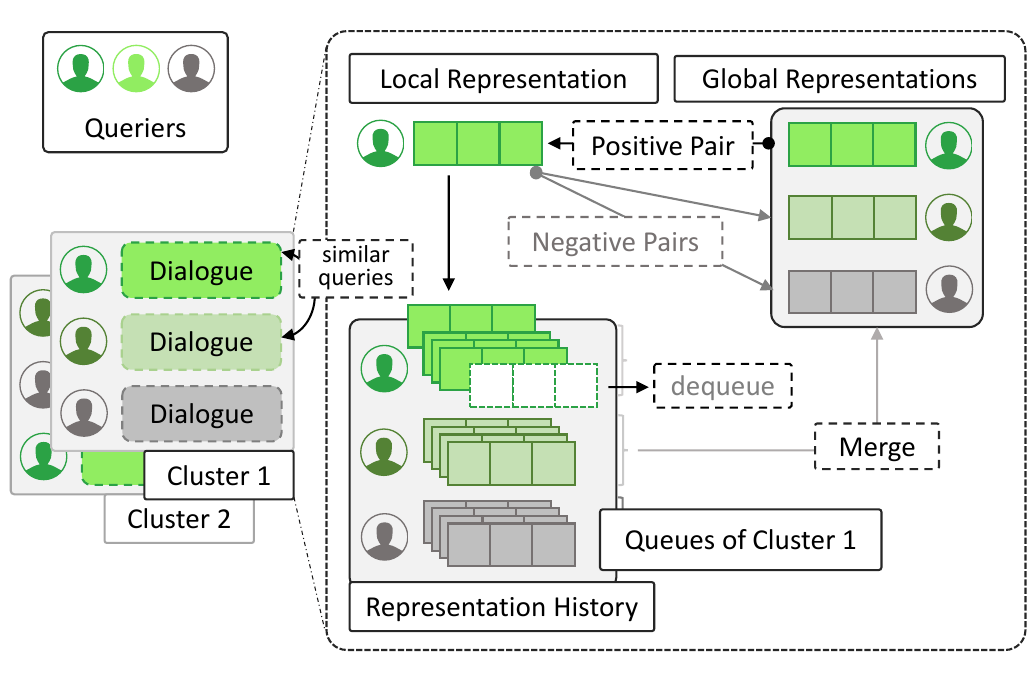}
    \caption{Querier-contrastive loss in each cluster of dialogues with similar queries from different queriers.}\label{fig:loss}
    \vspace{-0.6em}
\end{figure}

The goal of the querier-contrastive loss function is to pull close the representations of dialogues from the same querier, while pushing apart the representations of dialogues from different queriers. As shown in Figure \ref{fig:loss}, for a dialogue $d_k^{i}$ from $Q^{i}$ input into $G_s$, the output representation vector $G_s(x_k^{i})$ is projected by a learnable matrix $W$ to a low-dimensional space \cite{albef} for contrastive learning, denoted as
\begin{equation}
    z_k^{i} = W \cdot G_s(x_k^{i}).
\end{equation}
For each querier $Q^{i} \in \mathbb{Q}$, a global low-dimensional representation $e^{i}$ is maintained by averaging the representations of all $Q^{i}$'s previous dialogues during the training process. Then, $(z_k^{i}, e^{i})$ both from $Q^{i}$ constitute a positive pair, and $(z_k^{i}, e^{j})$ from $Q^{i}$ and $Q^{j}$ constitute a negative pair, where $e^{j}$ denotes global representation of $Q^{j}$. We further introduce a metric $f_{QC}(\cdot)$ to measure the distance between representations in each pair with a temperature hyperparameter $\tau$, formalized as 
\begin{equation}
    f_{QC}(z_k^{i}, e^{i}) = \exp\left(\frac{z_k^{i} \cdot e^{i}}{\Vert z_k^{i} \Vert \cdot \Vert e^{i} \Vert} \cdot \frac{1}{\tau}\right),
\end{equation} 
which is proportional to the cosine distance between normalized $z_j^i$ and $e^i$. We finally define the querier-contrastive loss as 
\begin{equation}\label{eq:loss:QC}
    L_{QC} = -\log \frac{f_{QC}(z_k^{i}, e^{i})}{\sum_{j=1}^m f_{QC}(z_k^{i}, e^{j})},
\end{equation}
where all $m$ queriers are introduced for contrast. In fact, the querier-contrastive loss can also be interpreted as a categorical loss that differentiates the dialogue features corresponding to different queriers. We can prove that minimizing this loss is to maximize the mutual information between the current dialogue representation and the corresponding querier's global representation. 
Specifically, referring to the main theorem in \cite{cpc}, we establish Lemma \ref{lemma1} as follow. 
\begin{lemma}\label{lemma1}
Given a feature vector $y$ and a set $X=\{x_1,\dots,x_n\}$ of $n$ features of random samples, where one sample $x_i$ is drawn from the conditional distribution $p(x_i|y)$ and the rest from the corresponding marginal distribution $p(x)$, an arbitrary score function $f$, and loss $L= -\mathop{\mathbb{E}}\limits_{X}\left[\log \frac{f(x_i, y)}{\sum_{x_j\in X}f(x_j, y)} \right]$, we have
\begin{equation}
    I(x_i, y) \geq \log n - L, 
\end{equation}
where $I(x_i, y)$ denotes the mutual information between $x_i$ and $y$. 
\end{lemma}

\begin{theorem} 
Given the representation $z_k^{i}$ of $d^{i}_k$ and the global feature set of queriers $E=\{e^{1},\dots, e^{m}\}$, minimizing the loss function $L_{QC}$ maximizes the mutual information between $d^{i}_k$ and $e^{i}$.
\end{theorem}

\begin{proof}
The loss $L$ in Lemma \ref{lemma1} corresponds to a categorical cross-entropy loss of classifying which $x_i$ is sampled from the conditional probability. Substituting $y$ with $z_k^{i}$, $X$ with $E$ and defining the score function $f(z_k^{i}, e^{i})=\exp(\frac{z_k^{i} \cdot e^{i}}{\tau \Vert z_k^{i} \Vert \cdot \Vert e^{i} \Vert})$ in Lemma \ref{lemma1}, we can find that $L_{QC}$ equals to $L$ in Lemma \ref{lemma1}. Then, we have 
\begin{equation}
    I(e^{i}, d^{i}_k) \geq \log m - L_{QC}. 
\end{equation}
Thus, minimizing $L_{QC}$ maximizes a lower bound on mutual information between $d^{i}_k$ and $e^{i}$. 
\end{proof}

\subsection{Multi-View Loss Augmentation}

We adopt the manual view augmentation technique \cite{clvr, learningsdg, multiviewinformation} to enhance querier-contrastive learning. In particular, we generate multiple views for each dialogue by applying different projection matrices. We let $W_{v_1}$ and $W_{v_2}$ denote the projection matrices for generating two views. Then, each dialogue $d_k^{i}$ has two-view representation vectors, denoted as $z_{k,v_1}^{i}$ and $z_{k,v_2}^{i}$. Similarly, each querier $Q_i$ also maintains two-view global representations, denoted as $e_{v_1}^{i}$ and $e_{v_2}^{i}$. The querier-contrastive loss with two-view augmentation becomes
\begin{equation}\label{eq:loss:qc:mv}
\begin{aligned}
        L_{QC}^{+} = -\frac{1}{2}\left(\log \frac{f(z_{k,{v_1}}^{i}, e^{i}_{v_2})}{\sum_{j=1}^m f(z_{k,{v_1}}^{i}, e^{j}_{v_2})}\right. \left. + \log \frac{f(z_{k,{v_2}}^{i}, e^{i}_{v_1})}{\sum_{j=1}^m f(z_{k,{v_2}}^{i}, e^{j}_{v_1})}\right),
\end{aligned}
\end{equation}
which contrasts the current dialogue representation with the global representation of the same querier across different views.

\section{Multi-Querier Dialogue (MQDialog) Dataset and Evaluation Metric}\label{sec:dataset}

\begin{table*}[!t]
\centering
\caption{MQDialog dataset statistics. It comprises a total of 173 queriers and 12 responders.}\label{table:dataset}
\vspace{-0.4em}
\resizebox{0.95\linewidth}{!}{
\begin{tabular}{lllllllll}
\toprule
Language               &  Data Source                           & \# Queriers & Querier Examples                          & Responder      & \# train & \# test & Avg. \# dialogues & Avg. \# turns\\
\midrule       
\multirow{6}*{English} & \multirow{2}*{The Big Bang Theory}     & 14          & Priya, Barry, Howard, Leonard, etc.       & Sheldon       & 4805 & 1101   & 333.79 & 3.29 \\ 
                       &                                        & 12          & Bernadette, Penny, Raj, Stuart, etc.      & Leonard       & 4607 & 1014   & 397.17 & 3.23 \\    
\cmidrule{2-9}
                       & \multirow{2}*{Friends}                 & 12          & Amy, Chandler, Charlie, Joey, etc.        & Rachel        & 3768 & 870    & 323.92 & 3.30 \\ 
                       &                                        & 20          & Ben, Mike, Gary, Paul, etc.               & Ross          & 3839 & 960    & 206.80 & 3.24 \\ 
                       \cmidrule{2-9}
                       & \multirow{2}*{Modern Family}           & 9           & Alex, Cameron, Dylan, Gloria, etc.        & Claire        & 1161 & 281    & 148.67 & 3.10 \\ 
                       &                                        & 8           & Haley, Jay, Luke, Mitchell, etc.          & Phil          & 881  & 246    & 136.12 & 3.03 \\ 
\midrule
\multirow{6}*{Chinese} & \multirow{3}*{My Own Swordsman}        & 16          & Bai Sanniang, Guo Furong, Mo Xiaobei, etc.& Tong Xiangyu  & 3200 & 831    & 219.44 & 3.52 \\ 
                       &                                        & 16          & Bao Daren, Ji Wuming, Zhu Wushuang, etc.  & Bai Zhantang  & 2995 & 857    & 217.88 & 3.38 \\ 
                       &                                        & 8           & Li Dazui, Xing Butou, Yan Xiaoliu, etc.   & Lv Xiucai     & 1635 & 409    & 229.50 & 3.39 \\ 
                       \cmidrule{2-9}
                       & \multirow{2}*{Empresses in the Palace} & 17          & Cao Guiren, Mei Zhuang, Liu Zhu, etc.     & Zhen Huan     & 1229 & 350    & 63.94  & 3.79 \\ 
                       &                                        & 11          & Consort Hua, Empress, Huan Bi, etc.       & Emperor       & 704  & 200    & 55.45  & 3.90  \\ 
                       \cmidrule{2-9}
                       & WeChat Records      & 30                             & Author's contacts                         & Author        & 16302 & 3887  & 71.53  & 16.67 \\ 
\bottomrule
\end{tabular}
}
\vspace{-0.2em}
\end{table*}

\subsection{Existing Datasets}
For querier-aware response evaluation, the desired dataset should include the dialogues between multiple queriers and one responder, where the responses should reflect the unique personality of each querier. Real-life private conversations from messaging applications, such as WhatsApp, WeChat, and Line, naturally exhibit these characteristics. However, due to privacy concerns, no public real-life conversation dataset is available. Other datasets or sources either lack clear differentiation between queriers (e.g., role-play datasets \cite{characterllm,chatharuhi} that feature one user player interacting with independent characters and ShareGPT dataset with anonymous, non-identifiable queriers \cite{sharegpt}) or contain too few dialogues between each querier-responder pair (e.g., Charactereval \cite{charactereval}, Reditt, and Quora).
Thus, we build a new multi-querier dialogue dataset, called MQDialog, sourced from English and Chinese scripts and real-world WeChat records. 
In what follows, we introduce data sources, processing methods, statistics, and evaluation metrics. 

\subsection{Data Sources}

We take a diverse set of source corpora, including 3 English scripts, 2 Chinese scripts, and 1 person’s WeChat records. We select 12 leading actors as responders based on the volume of dialogue they contribute, including Sheldon Cooper and Leonard Hofstadter from {\em The Big Bang Theory}, Rachel Green and Ross Geller from {\em Friends}, Claire and Phil Dunphy from {\em Modern Family}, Tong Xiangyu, Bai Zhantang, and Lv Xiucai from {\em My Own Swordsman}, Zhen Huan and the Emperor from {\em Empresses in the Palace}, and one {\em WeChat user} who is the author of the work. All scripts are collected from the internet and the source links are presented in Appendix \ref{sec:data_source_link}. For each leading actor acting as the responder, the other supporting actors interacting with them are considered distinct queriers.
As for the WeChat dataset, the author serves as the responder, and the queriers include labmates, professors from the same lab, and colleagues from a partner organization. They also agree that the dataset can be properly used for research purposes. Since there are many minor characters in the scripts who appear infrequently, we set a threshold of 20 dialogues with the responder for each querier, which effectively filters out most of them.

\subsection{Dialogue Extraction and Cleaning} \label{sec:data_construct}
For scripts, raw data consists of multi-person conversations, and we extract the dialogues between each querier-responder pair of a supporting actor and a leading actor. We retain the interactions within an episode as a multi-turn conversation. For each conversation, we remove any content preceding the supporting actor’s first line, ensuring that the supporting actor strictly functions as the querier, while the leading actor serves as the responder. Additionally, to improve data quality, we eliminate errors commonly found in the original scripts sourced from the internet, such as repeated conversations or blank entries. For WeChat records, we focus on two-person dialogues and exclude group chats. A new dialogue begins when the current message is sent after a predefined interval (e.g., empirically set to 3 hours in this work) following the previous message, signifying the start of a multi-turn conversation.

\subsection{Dataset Statistics and Splitting}

The MQDialog dataset has 173 queriers and 12 responders in total. We randomly split the dialogues into the training and test sets for each querier, and then aggregate these sets based on the responder. On average, each responder has 3761 dialogues with different queriers for training and 917 dialogues for testing. 
Additionally,  the average number of dialogues per querier ranges from 55.45 to 397.17. For datasets sourced from scripts, the average number of turns per dialogue falls between 3 and 4, whereas real-world WeChat conversations have an average of 16.67 turns per dialogue. The detailed statistics are shown in Table \ref{table:dataset}.

\begin{table*}[!t]
\centering
\caption{Our design vs. Baselines of zero-shot generation and finetuning from BLEU and ROUGE (\%) metrics, where "$*$" denotes the result is significantly worse than our method in t-test with $p < 0.05$ level.}\label{main_result}
\vspace{-0.4em}
\resizebox{0.97\linewidth}{!}{
\begin{tabular}{lrrrrrrrrrrrr}
    \toprule
        \multirow{2}{*}{Responder} &  \multicolumn{3}{c}{BLEU} & \multicolumn{3}{c}{ROUGE-1}  & \multicolumn{3}{c}{ROUGE-2}  & \multicolumn{3}{c}{ROUGE-L} \\ 
        \cmidrule{2-13}
        ~ & \makecell[c]{Zero-shot} & \makecell[c]{FT} & \makecell[c]{Ours} & \makecell[c]{Zero-shot} & \makecell[c]{FT} & \makecell[c]{Ours} & \makecell[c]{Zero-shot} & \makecell[c]{FT} & \makecell[c]{Ours} & \makecell[c]{Zero-shot} & \makecell[c]{FT} & \makecell[c]{Ours} \\ 
        \midrule
        Sheldon	&	$3.98_{\pm 0.04}$	&	$6.57_{\pm 0.08}$	&	$\textbf{7.16}_{\pm 0.09}$	&	$6.18_{\pm 0.08}$	&	$6.67_{\pm 0.12}^*$	&	$\textbf{7.52}_{\pm 0.11}$	&	$0.47_{\pm 0.03}$	&	$0.98_{\pm 0.06}$	&	$\textbf{1.00}_{\pm 0.07}$	&	$7.21_{\pm 0.08}$	&	$6.67_{\pm 0.12}^*$	&	$\textbf{7.40}_{\pm 0.11}$	\\
        Leonard	&	$2.56_{\pm 0.03}^*$	&	$5.98_{\pm 0.18}^*$	&	$\textbf{6.14}_{\pm 0.18}$	&	$4.38_{\pm 0.12}$	&	$7.72_{\pm 0.19}^*$	&	$\textbf{8.26}_{\pm 0.19}$	&	$0.37_{\pm 0.06}$	&	$1.50_{\pm 0.10}$	&	$\textbf{1.71}_{\pm 0.11}$	&	$5.49_{\pm 0.12}$	&	$7.65_{\pm 0.19}^*$	&	$\textbf{8.21}_{\pm 0.19}$	\\
        Rachel	&	$2.92_{\pm 0.03}^*$	&	$6.52_{\pm 0.13}$	&	$\textbf{6.69}_{\pm 0.15}$	&	$4.69_{\pm 0.10}$	&	$6.45_{\pm 0.15}$	&	$\textbf{7.12}_{\pm 0.16}$	&	$0.35_{\pm 0.05}$	&	$1.07_{\pm 0.08}$	&	$\textbf{1.16}_{\pm 0.08}$	&	$5.55_{\pm 0.10}^*$	&	$6.48_{\pm 0.15}$	&	$\textbf{7.18}_{\pm 0.16}$	\\
        Ross	&	$3.11_{\pm 0.03}$	&	$5.89_{\pm 0.14}$	&	$\textbf{6.34}_{\pm 0.15}$	&	$4.88_{\pm 0.09}$	&	$6.29_{\pm 0.15}$	&	$\textbf{7.02}_{\pm 0.16}$	&	$0.41_{\pm 0.04}$	&	$1.07_{\pm 0.07}$	&	$\textbf{1.08}_{\pm 0.07}$	&	$5.80_{\pm 0.09}$	&	$6.33_{\pm 0.15}$	&	$\textbf{7.06}_{\pm 0.16}$	\\
        Claire	&	$2.56_{\pm 0.03}^*$	&	$5.51_{\pm 0.04}$	&	$\textbf{5.76}_{\pm 0.12}$	&	$4.30_{\pm 0.09}$	&	$5.10_{\pm 0.10}$	&	$\textbf{6.04}_{\pm 0.14}$	&	$0.28_{\pm 0.06}$	&	$0.76_{\pm 0.05}$	&	$\textbf{0.83}_{\pm 0.06}$	&	$5.21_{\pm 0.09}$	&	$5.67_{\pm 0.10}$	&	$\textbf{6.16}_{\pm 0.14}$	\\
        Phil	&	$2.74_{\pm 0.03}$	&	$4.88_{\pm 0.04}$	&	$\textbf{5.71}_{\pm 0.08}$	&	$4.91_{\pm 0.07}$	&	$5.68_{\pm 0.09}$	&	$\textbf{5.75}_{\pm 0.11}$	&	$0.44_{\pm 0.03}$	&	$0.85_{\pm 0.06}$	&	$\textbf{0.99}_{\pm 0.07}$	&	$6.14_{\pm 0.07}^*$	&	$6.22_{\pm 0.09}$	&	$\textbf{7.24}_{\pm 0.08}$	\\
        \midrule
        Avg.    & 2.9789 & 5.8924 &	\textbf{6.3000} & 4.8927	& 6.3194 & \textbf{6.9514} &	0.3878 & 1.0377	& \textbf{1.1266} & 5.9002 &	6.5032 & \textbf{7.2090} \\
        \toprule
        Tong Xiangyu	&	$3.35_{\pm 0.06}^*$	&	$8.89_{\pm 0.11}$	&	$\textbf{9.42}_{\pm 0.11}$	&	$10.8_{\pm 0.12}^*$	&	$14.82_{\pm 0.14}^*$	&	$\textbf{16.03}_{\pm 0.14}$	&	$1.40_{\pm 0.06}^*$	&	$3.00_{\pm 0.10}$	&	$\textbf{3.20}_{\pm 0.11}$	&	$10.26_{\pm 0.11}^*$	&	$14.54_{\pm 0.14}^*$	&	$\textbf{15.50}_{\pm 0.14}$	\\
        Bai Zhantang	&	$3.43_{\pm 0.08}^*$	&	$9.03_{\pm 0.15}$	&	$\textbf{9.87}_{\pm 0.14}$	&	$10.99_{\pm 0.14}^*$	&	$16.00_{\pm 0.16}^*$	&	$\textbf{17.26}_{\pm 0.17}$	&	$1.79_{\pm 0.09}^*$	&	$3.68_{\pm 0.14}^*$	&	$\textbf{4.14}_{\pm 0.14}$	&	$10.57_{\pm 0.14}^*$	&	$14.78_{\pm 0.16}^*$	&	$\textbf{15.84}_{\pm 0.18}$	\\
        Lv Xiucai	&	$3.28_{\pm 0.06}^*$	&	$7.85_{\pm 0.15}$	&	$\textbf{9.19}_{\pm 0.15}$	&	$10.16_{\pm 0.12}^*$	&	$13.90_{\pm 0.13}$	&	$\textbf{14.65}_{\pm 0.16}$	&	$1.23_{\pm 0.07}^*$	&	$3.40_{\pm 0.13}$	&	$\textbf{3.98}_{\pm 0.14}$	&	$10.25_{\pm 0.12}^*$	&	$13.99_{\pm 0.13}$	&	$\textbf{14.82}_{\pm 0.16}$	\\
        Zhen Huan	&	$4.34_{\pm 0.08}^*$	&	$9.02_{\pm 0.15}$	&	$\textbf{10.02}_{\pm 0.13}$	&	$14.58_{\pm 0.13}^*$	&	$21.48_{\pm 0.17}$	&	$\textbf{22.57}_{\pm 0.15}$	&	$3.03_{\pm 0.09}^*$	&	$6.33_{\pm 0.15}$	&	$\textbf{6.74}_{\pm 0.15}$	&	$12.87_{\pm 0.13}^*$	&	$18.95_{\pm 0.16}^*$	&	$\textbf{20.18}_{\pm 0.15}$	\\
        Emperor	&	$4.17_{\pm 0.06}^*$	&	$10.74_{\pm 0.12}$	&	$\textbf{10.78}_{\pm 0.16}$	&	$13.14_{\pm 0.14}$	&	$18.14_{\pm 0.16}$	&	$\textbf{18.75}_{\pm 0.14}$	&	$1.65_{\pm 0.13}$	&	$3.97_{\pm 0.14}$	&	$\textbf{4.40}_{\pm 0.14}$	&	$11.58_{\pm 0.14}^*$	&	$16.43_{\pm 0.15}$	&	$\textbf{17.44}_{\pm 0.14}$	\\
        Real Person	&	$5.90_{\pm 0.14}$	&	$8.19_{\pm 0.24}^*$	&	$\textbf{9.96}_{\pm 0.22}$	&	$11.58_{\pm 0.18}^*$	&	$17.79_{\pm 0.29}^*$	&	$\textbf{18.20}_{\pm 0.28}$	&	$4.28_{\pm 0.15}^*$	&	$10.87_{\pm 0.27}$	&	$\textbf{11.02}_{\pm 0.26}$	&	$11.87_{\pm 0.19}^*$	&	$18.65_{\pm 0.30}^*$	&	$\textbf{19.16}_{\pm 0.29}$	\\
        \midrule
        Avg.    & 4.0765	&	8.9543	&	\textbf{9.8709}	&	11.8744	&	17.0210	&	\textbf{17.9098}	&	2.2297	&	5.2084	&	\textbf{5.5811}	&	11.2336	&	16.2228	&	\textbf{17.1562} \\
        \bottomrule
    \end{tabular}
}
\vspace{-0.4em}
\end{table*}

\subsection{Evaluation Methods and Metrics} \label{sec:dataset_metric}
To evaluate the quality of response generation, we adopt two kinds of automatic evaluation methods. (1) The first method is comparing generated responses with ground-truth references by quantitatively measuring their lexical and semantic similarity. 
Since the ground-truth responses provided by responders in the dataset inherently reflect personalization tailored to different queriers' needs and preferences, we can evaluate personalization by assessing the consistency between the predicted responses and the ground truth. The closer the predicted response is to the ground truth, the better it captures personalized information.
We take widely used metrics, including BLEU \cite{bleu} and ROUGE variants \cite{rouge} (i.e., ROUGE-1, ROUGE-2, and ROUGE-L). (2) The second method is to compare the responses generated by the proposed design and the baseline using a powerful LLM (e.g., GPT-4-turbo-2024-04-09\footnote{https://openai.com/index/gpt-4/} in this work), which is popular in recent work \cite{chatgptoutperform, geval}. We record the winning rate as the evaluation metric. In particular, we prompt LLM to judge which of the two responses is more like the words the responder would speak to a specific querier, ensuring a closer alignment with the personalities of the querier and the responder, as well as their relationship. During testing, we randomly sample 5 dialogues between the pair of the querier and the responder from the training set as few-shot examples for prompting the LLM evaluator. 
We reserve the detailed prompts in Appendix \ref{sec:eval_prompt}.

\section{Evaluation}

\begin{table*}[!ht]
\centering
\caption{Our design vs. Baselines of responder profile-based generation, querier characteristic-based generation, and few-shot generation from BLEU and ROUGE (\%) metrics, where "$*$" denotes the result is significantly worse than our method in t-test.}\label{table:rgp_full_result}
\vspace{-0.4em}
\resizebox{0.97\linewidth}{!}{
    \begin{tabular}{lrrrrrrrrrrrr}
    \toprule
        \multirow{2}{*}{Responder} & \multicolumn{4}{c}{BLEU} & \multicolumn{4}{c}{ROUGE-1} & \multicolumn{4}{c}{ROUGE-L} \\ 
        \cmidrule{2-13}
        ~ & \makecell[c]{RPG} & \makecell[c]{Few-shot} & \makecell[c]{QCG} & \makecell[c]{Ours} & \makecell[c]{RPG} & \makecell[c]{Few-shot} & \makecell[c]{QCG} & \makecell[c]{Ours} & \makecell[c]{RPG} & \makecell[c]{Few-shot} & \makecell[c]{QCG} & \makecell[c]{Ours}  \\ 
        \midrule
Sheldon	&	$3.78_{\pm 0.04}^*$	&	$4.13_{\pm 0.04}$	&	$5.08_{\pm 0.06}^*$	&	$\textbf{7.16}_{\pm 0.09}$	&	$6.11_{\pm 0.07}$	&	$6.13_{\pm 0.08}$	&	$7.44_{\pm 0.08}$	&	$\textbf{7.52}_{\pm 0.11}$	&	$6.82_{\pm 0.07}$	&	$7.09_{\pm 0.08}$	&	$7.08_{\pm 0.07}$	&	$\textbf{7.40}_{\pm 0.11}$	\\
Leonard	&	$2.41_{\pm 0.03}^*$	&	$2.56_{\pm 0.03}^*$	&	$3.87_{\pm 0.06}$	&	$\textbf{6.14}_{\pm 0.18}$	&	$4.21_{\pm 0.07}^*$	&	$4.43_{\pm 0.13}^*$	&	$5.73_{\pm 0.08}$	&	$\textbf{8.26}_{\pm 0.19}$	&	$4.75_{\pm 0.07}^*$	&	$5.34_{\pm 0.13}^*$	&	$5.44_{\pm 0.08}^*$	&	$\textbf{8.21}_{\pm 0.19}$	\\
Rachel	&	$2.77_{\pm 0.03}^*$	&	$2.87_{\pm 0.04}^*$	&	$3.81_{\pm 0.06}$	&	$\textbf{6.69}_{\pm 0.15}$	&	$4.62_{\pm 0.08}$	&	$4.51_{\pm 0.09}^*$	&	$5.59_{\pm 0.08}$	&	$\textbf{7.12}_{\pm 0.16}$	&	$5.08_{\pm 0.08}$	&	$5.18_{\pm 0.09}^*$	&	$5.19_{\pm 0.07}^*$	&	$\textbf{7.18}_{\pm 0.16}$	\\
Ross	&	$2.87_{\pm 0.03}$	&	$3.08_{\pm 0.04}$	&	$3.71_{\pm 0.05}$	&	$\textbf{6.34}_{\pm 0.15}$	&	$4.75_{\pm 0.07}$	&	$4.80_{\pm 0.09}^*$	&	$5.82_{\pm 0.07}$	&	$\textbf{7.02}_{\pm 0.16}$	&	$5.30_{\pm 0.07}$	&	$5.33_{\pm 0.09}$	&	$5.45_{\pm 0.06}$	&	$\textbf{7.06}_{\pm 0.16}$	\\
Claire	&	$2.31_{\pm 0.03}^*$	&	$2.59_{\pm 0.03}$	&	$2.90_{\pm 0.05}$	&	$\textbf{5.76}_{\pm 0.12}$	&	$4.14_{\pm 0.07}$	&	$4.26_{\pm 0.09}$	&	$5.02_{\pm 0.07}$	&	$\textbf{6.04}_{\pm 0.14}$	&	$4.83_{\pm 0.07}$	&	$5.21_{\pm 0.10}$	&	$4.78_{\pm 0.06}$	&	$\textbf{6.16}_{\pm 0.14}$	\\
Phil	&	$2.37_{\pm 0.03}^*$	&	$2.42_{\pm 0.03}$	&	$3.14_{\pm 0.04}$	&	$\textbf{5.71}_{\pm 0.08}$	&	$4.28_{\pm 0.05}$	&	$4.27_{\pm 0.07}$	&	$4.42_{\pm 0.06}$	&	$\textbf{5.75}_{\pm 0.11}$	&	$4.66_{\pm 0.05}^*$	&	$5.07_{\pm 0.07}^*$	&	$4.18_{\pm 0.06}^*$	&	$\textbf{7.24}_{\pm 0.08}$	\\
\midrule
Avg.	&	2.7512	&	2.9411	&	3.7517	&	\textbf{6.3000}	&	4.6860	&	4.7314	&	5.6700	&	\textbf{6.9514}	&	5.2393	&	5.5360	&	5.3533	&	\textbf{7.2090}	\\
\toprule
Tong Xiangyu	&	$4.33_{\pm 0.05}^*$	&	$6.04_{\pm 0.09}^*$	&	$5.80_{\pm 0.06}^*$	&	$\textbf{9.42}_{\pm 0.11}$	&	$11.49_{\pm 0.08}^*$	&	$10.35_{\pm 0.12}^*$	&	$14.16_{\pm 0.09}^*$	&	$\textbf{16.03}_{\pm 0.14}$	&	$9.90_{\pm 0.06}^*$	&	$9.52_{\pm 0.11}^*$	&	$11.07_{\pm 0.07}^*$	&	$\textbf{15.50}_{\pm 0.14}$	\\
Bai Zhantang	&	$4.42_{\pm 0.05}^*$	&	$5.93_{\pm 0.08}^*$	&	$5.37_{\pm 0.06}^*$	&	$\textbf{9.87}_{\pm 0.14}$	&	$11.12_{\pm 0.09}^*$	&	$9.35_{\pm 0.12}^*$	&	$13.05_{\pm 0.10}^*$	&	$\textbf{17.26}_{\pm 0.17}$	&	$9.74_{\pm 0.07}^*$	&	$8.82_{\pm 0.11}^*$	&	$10.51_{\pm 0.08}^*$	&	$\textbf{15.84}_{\pm 0.18}$	\\
Lv Xiucai	&	$4.21_{\pm 0.05}^*$	&	$3.01_{\pm 0.08}^*$	&	$5.21_{\pm 0.07}^*$	&	$\textbf{9.19}_{\pm 0.15}$	&	$10.80_{\pm 0.09}^*$	&	$6.80_{\pm 0.14}$	&	$12.60_{\pm 0.09}^*$	&	$\textbf{14.65}_{\pm 0.16}$	&	$9.54_{\pm 0.08}^*$	&	$6.37_{\pm 0.13}^*$	&	$10.67_{\pm 0.08}^*$	&	$\textbf{14.82}_{\pm 0.16}$	\\
Zhen Huan	&	$6.06_{\pm 0.05}^*$	&	$9.91_{\pm 0.11}$	&	$7.46_{\pm 0.07}^*$	&	$\textbf{10.02}_{\pm 0.13}$	&	$16.43_{\pm 0.09}^*$	&	$19.06_{\pm 0.13}$	&	$18.38_{\pm 0.09}^*$	&	$\textbf{22.57}_{\pm 0.15}$	&	$13.01_{\pm 0.07}^*$	&	$16.84_{\pm 0.12}^*$	&	$14.58_{\pm 0.08}^*$	&	$\textbf{20.18}_{\pm 0.15}$	\\
Emperor	&	$5.82_{\pm 0.06}^*$	&	$8.72_{\pm 0.1}$	&	$9.05_{\pm 0.07}$	&	$\textbf{10.78}_{\pm 0.16}$	&	$14.83_{\pm 0.09}^*$	&	$16.69_{\pm 0.14}$	&	$18.10_{\pm 0.09}$	&	$\textbf{18.75}_{\pm 0.14}$	&	$11.54_{\pm 0.07}^*$	&	$14.63_{\pm 0.13}$	&	$14.60_{\pm 0.08}^*$	&	$\textbf{17.44}_{\pm 0.14}$	\\
Real Person	&	-	&	$2.12_{\pm 0.18}^*$	&	-	&	$\textbf{9.96}_{\pm 0.22}$	&	-	&	$12.93_{\pm 0.22}^*$	&	-	&	$\textbf{18.20}_{\pm 0.28}$	&	-	&	$13.31_{\pm 0.23}^*$	&	-	&	$\textbf{19.16}_{\pm 0.29}$	\\

\midrule
Avg.	&	4.9687	&	5.9532	&	6.5780	&	\textbf{9.8709}	&	12.9347	&	12.5298	&	15.2580	&	\textbf{17.9098}	&	10.7451	&	11.5823	&	12.2860	&	\textbf{17.1562}	\\
        \bottomrule
    \end{tabular}
}
\vspace{-0.4em}
\end{table*}

\subsection{Experimental Setups}

\textbf{Pretrained Models.} 
We take the chat version of different pretrained LLMs. We use Llama3-8B-Instruction \cite{llama} for datasets in English and Qwen1.5-7B-Chat \cite{qwen} for datasets in Chinese. All the pretrained checkpoints are loaded from \href{https://huggingface.co}{huggingface}\footnote{https://huggingface.co}.

\textbf{Baselines.} We introduce the following baselines for comparison: 
(1) {\bf Zero-shot generation} \cite{gpt3}, which directly answers the query with the pretrained LLM;
(2) {\bf Finetuning (FT)}, which tunes the pretrained LLM over the whole training set with the mixed dialogues between the responder and different queriers;
(3) {\bf Responder profile-based generation (RPG)} \cite{harrypotter, rolellm}, which concatenates the manually constructed role profile of the responder and the query together as LLM input;
(4) {\bf Few-shot prompting generation} \cite{icl}, which retrieves five dialogues from the training set as examples and concatenates them with the current query based on a template as LLM input;
and (5) {\bf Querier characteristic-based generation (QCG)}, which concatenates the constructed querier's characteristic prompt and the query together as the input. Specifically, since there are more than 100 queriers, we leverage GPT-4o to automatically generate each querier's characteristic prompt, including profile and relationship with the responder, using the instructions shown in Table \ref{table:instruction_for_qpg}. Then we manually review the characteristic prompts to ensure correctness and semantic completeness. 

To validate the effectiveness of the designed loss and clustering strategy, we further introduce the following variants of our design: (6) {\bf Querier-aware responder without querier-contrastive loss (w/o QCL)}, which uses additional parameters for encoding specific personality without the supervision of querier-contrastive loss; and
(7) {\bf Querier-aware responder without cluster-based contrastive learning (w/o CCL)}, which performs contrastive learning over all the dialogues without clustering strategy rather than each cluster with similar queries.

\begin{table}[!t]
\small
\centering
\caption{Instructions for generating querier's characteristic. }\label{table:instruction_for_qpg}
\resizebox{0.95\columnwidth}{!}{
    \centering
    \begin{tabular}{l|p{0.85\linewidth}}
    \toprule
Language & Instructions \\
\midrule
English & Provide a brief introduction to <QUERIER> in <DATASET-NAME> and describe his/her relationship with <RESPONDER> in one short paragraph. \\
\midrule
Chinese & \chinese{用一段话简短介绍<DATASET-NAME>中的<QUERIER>，以及他/她跟<RESPONDER>之间的关系。} \\
    \bottomrule
    \end{tabular}
    }
\end{table}

\textbf{Implementation Details.} We implement our design and all the baselines in PyTorch. The workstation has 8 NVIDIA V100 32G GPUs. To improve training efficiency, we integrate LoRA \cite{lora} for the general block. Additionally, we also share parameters between the attention layers of the transformer blocks in the general encoder and the corresponding layers in the specific encoder. The number of dialogue clusters is set to 10, resulting in an average within-cluster query similarity of 0.83 across all datasets.
We employ the AdamW optimization scheme for all methods. The maximum learning rate is set to 2e-4, the minimum learning rate is set to 1e-4, the batch size is set to 4, the epoch number is set to 20, and the maximum length of tokens is set to 592. For LoRA, we set the rank to 16, the dropout rate to 0.05, and apply weight decomposition for the query, key, value matrices, and the dense MLP in a transformer block. The parameters of models are quantized to FP16 during training.

\subsection{From BLEU and ROUGE Metrics}

We first report the comparison results of our design with zero-shot generation and FT in Table \ref{main_result}. We can observe that, on average across all the responders, the relative improvements over zero-shot generation are 111.4\%, 42.1\%,  190.5\%, and 22.2\% for BLEU, ROUGE-1, ROUGE-2, and ROUGE-L, respectively, on the English dataset; and 142.1\%, 50.8\%, 150.3\%, and 52.7\% on the Chinese dataset.  We can also see that the average relative improvements over FT are 6.9\%, 10.0\%,  8.6\%, and 10.9\% for BLEU, ROUGE-1, ROUGE-2, and ROUGE-L, respectively, on the English dataset; and 10.2\%, 5.2\%, 7.2\%, and 5.8\% on the Chinese dataset. The significant improvements highlight the importance of considering the querier's personality to enhance the quality of generated responses.

We then report the performance of RPG, QCG, and few-shot generation baselines in Table \ref{table:rgp_full_result}. Compared to RPG, QCG gains average relative BLEU improvements of 36.3\% and 32.4\% on the English and Chinese datasets, respectively, and ROUGE-L improvements of 2.2\% and 14.3\%. These results suggest that accounting for the querier's personality and the relationship between the querier and responder is essential for generating effective personalized responses.
We then compare our design with these baselines. We can observe that, compared with RPG, the average ROUGE-L improvements are 37.6\% and 59.7\% on the English and Chinese datasets, respectively; compared with few-shot generation, the average ROUGE-L improvements are 30.2\% and 48.1\%; and compared with QCG, the average ROUGE-L improvements are 34.7\% and 39.6\%. These results highlight the importance of incorporating comprehensive interactions between multiple queriers and the responder into LLM.

\subsection{From Winning Rate Judged by GPT-4}\label{sec:win}

\begin{figure}[!t]
  \centering
    \begin{subfigure}{0.97\linewidth}
         \centering
         \includegraphics[width=0.48\textwidth]{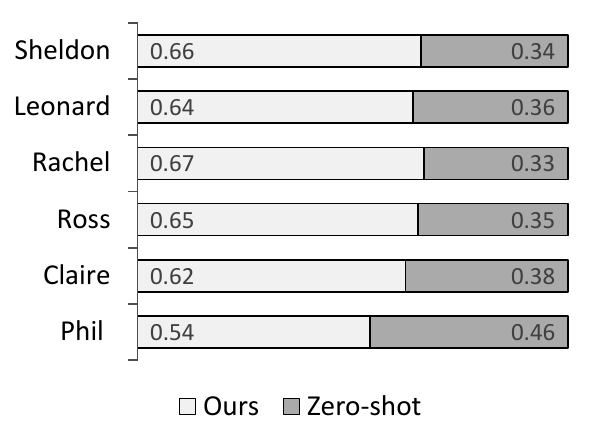}
         \includegraphics[width=0.48\textwidth]{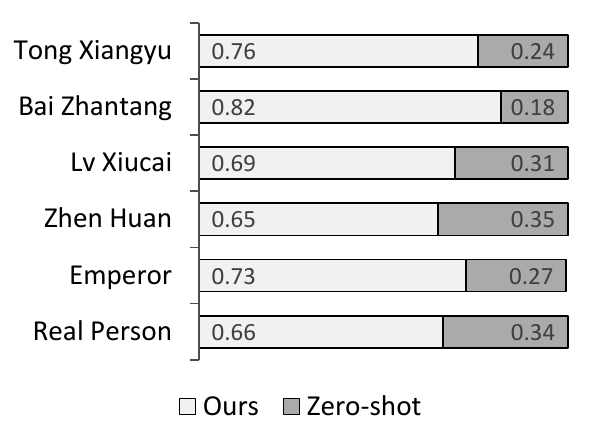}
         \caption{\centering Ours vs. Zero-Shot Generation}
         \label{fig:winrate_zg}
    \end{subfigure}
    \begin{subfigure}{0.97\linewidth}
         \centering
         \includegraphics[width=0.48\textwidth]{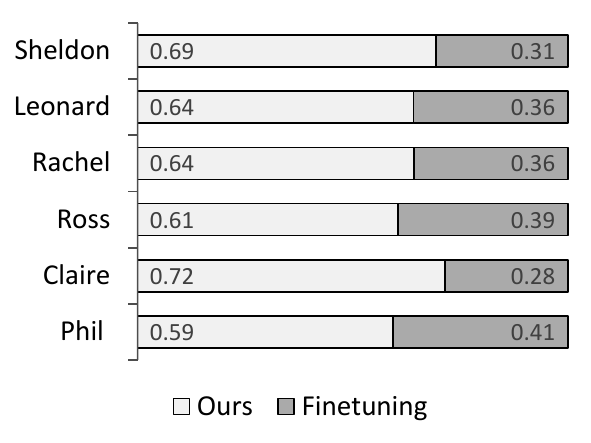}
         \includegraphics[width=0.48\textwidth]{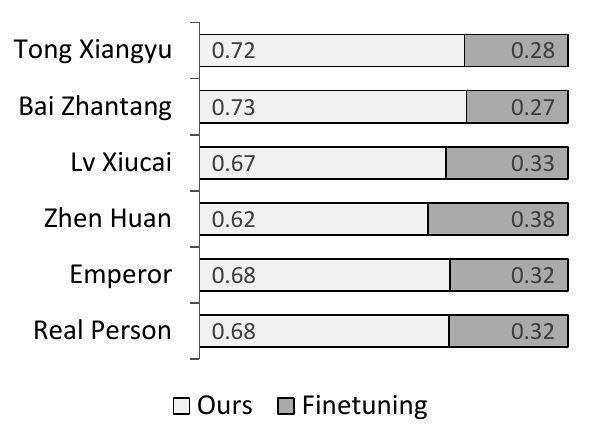}
         \caption{\centering Ours vs. Finetuning}
         \label{fig:winrate_ft}
    \end{subfigure}
    \vspace{-0.4em}
  \caption{The winning rate judged by GPT-4.}\label{fig:winrate}
  \vspace{-0.4em}
\end{figure}

The winning rates for our design in comparison to zero-shot generation and FT are shown in Figure \ref{fig:winrate}. The results are consistent with the findings from the BLEU and ROUGE metrics. In particular, compared with zero-shot generation, our design achieves winning rates ranging from 54\% to 67\% on the English dataset, and from 66\% to 82\% on the Chinese dataset. Against FT, the winning rates range from 59\% to 72\% on the English dataset, and from 62\% to 73\% on the Chinese dataset.

\textbf{Human Evaluation.} 
We randomly select 120 dialogues from 12 different responders, with 60 from the English portion and 60 from the Chinese portion of the dataset. The responses generated by our model and the baselines are then manually compared for quality and for which is more like what the responder would naturally say to the specific querier by annotators. The annotators are three volunteers in the lab. To maintain objectivity, annotators are blind to the method used to generate each response. Each annotator reviews 40 dialogues and their annotations are subsequently cross-checked by another annotator to ensure annotation quality and consistency.
The human evaluation results show that, compared to the zero-shot generation, our model achieves a winning rate of 74.2\%, and compared to FT, the winning rate is 67.5\%. These results demonstrate that the GPT-4 evaluation aligns with the manual evaluation outcomes.

\begin{table*}[!ht]
\centering
\caption{Ablation study results of querier-contrastive loss or query similarity-based clustering from BLEU and ROUGE metrics (\%), where "$*$" denotes the result is significantly worse than our method in t-test with $p < 0.05$ level.}\label{table:ablation_full_result}
\vspace{-0.4em}
\resizebox{0.97\linewidth}{!}{
    \begin{tabular}{lrrrrrrrrrrrr}
    \toprule
         \multirow{2}{*}{Responder} & \multicolumn{3}{c}{BLEU} & \multicolumn{3}{c}{ROUGE-1} & \multicolumn{3}{c}{ROUGE-2} & \multicolumn{3}{c}{ROUGE-L} \\ 
         \cmidrule{2-13}
        ~ & \makecell[c]{w/o QCL} & \makecell[c]{w/o CCL} & \makecell[c]{Ours} & \makecell[c]{w/o QCL} & \makecell[c]{w/o CCL} & \makecell[c]{Ours} & \makecell[c]{w/o QCL} & \makecell[c]{w/o CCL} & \makecell[c]{Ours} & \makecell[c]{w/o QCL} & \makecell[c]{w/o CCL} & \makecell[c]{Ours} \\ 
        \midrule
Sheldon	&	$6.69_{\pm 0.09}$	&	$6.92_{\pm 0.09}$	&	$\textbf{7.16}_{\pm 0.09}$	&	$7.19_{\pm 0.11}$	&	$7.16_{\pm 0.11}$	&	$\textbf{7.52}_{\pm 0.11}$	&	$\textbf{1.17}_{\pm 0.07}$	&	$1.04_{\pm 0.06}$	&	$1.00_{\pm 0.07}$	&	$7.04_{\pm 0.11}$	&	$7.14_{\pm 0.11}$	&	$\textbf{7.40}_{\pm 0.11}$	\\
Leonard	&	$6.14_{\pm 0.17}$	&	$5.90_{\pm 0.18}^*$	&	$\textbf{6.14}_{\pm 0.18}$	&	$7.40_{\pm 0.19}^*$	&	$7.77_{\pm 0.19}^*$	&	$\textbf{8.26}_{\pm 0.19}$	&	$1.68_{\pm 0.11}$	&	$1.49_{\pm 0.10}$	&	$\textbf{1.71}_{\pm 0.11}$	&	$7.38_{\pm 0.19}^*$	&	$7.78_{\pm 0.19}^*$	&	$\textbf{8.21}_{\pm 0.19}$	\\
Rachel	&	$6.34_{\pm 0.14}$	&	$5.79_{\pm 0.14}$	&	$\textbf{6.69}_{\pm 0.15}$	&	$6.69_{\pm 0.14}$	&	$6.83_{\pm 0.16}$	&	$\textbf{7.12}_{\pm 0.16}$	&	$0.98_{\pm 0.07}$	&	$1.00_{\pm 0.07}$	&	$\textbf{1.16}_{\pm 0.08}$	&	$6.87_{\pm 0.14}$	&	$6.96_{\pm 0.16}$	&	$\textbf{7.18}_{\pm 0.16}$	\\
Ross	&	$5.87_{\pm 0.15}$	&	$5.90_{\pm 0.13}$	&	$\textbf{6.34}_{\pm 0.15}$	&	$6.55_{\pm 0.16}$	&	$6.51_{\pm 0.14}$	&	$\textbf{7.02}_{\pm 0.16}$	&	$\textbf{1.15}_{\pm 0.08}$	&	$1.13_{\pm 0.08}$	&	$1.08_{\pm 0.07}$	&	$6.58_{\pm 0.16}$	&	$6.62_{\pm 0.14}$	&	$\textbf{7.06}_{\pm 0.16}$	\\
Claire	&	$5.53_{\pm 0.11}$	&	$5.25_{\pm 0.08}$	&	$\textbf{5.76}_{\pm 0.12}$	&	$5.58_{\pm 0.13}$	&	$5.45_{\pm 0.09}$	&	$\textbf{6.04}_{\pm 0.14}$	&	$0.58_{\pm 0.04}$	&	$0.76_{\pm 0.06}$	&	$\textbf{0.83}_{\pm 0.06}$	&	$5.53_{\pm 0.12}$	&	$6.08_{\pm 0.09}$	&	$\textbf{6.16}_{\pm 0.14}$	\\
Phil	&	$5.23_{\pm 0.10}$	&	$5.50_{\pm 0.12}$	&	$\textbf{5.71}_{\pm 0.08}$	&	$5.57_{\pm 0.11}$	&	$5.44_{\pm 0.13}$	&	$\textbf{5.75}_{\pm 0.11}$	&	$\textbf{1.10}_{\pm 0.07}$	&	$0.99_{\pm 0.07}$	&	$0.99_{\pm 0.07}$	&	$5.97_{\pm 0.10}$	&	$6.89_{\pm 0.14}$	&	$\textbf{7.24}_{\pm 0.08}$	\\
        \midrule
        Avg.    & 5.9665	&	5.8777	&	\textbf{6.3000}	&	6.4963	&	6.5283	&	\textbf{6.9514}	&	1.1087	&	1.0668	&	\textbf{1.1266}	&	6.5617	&	6.9115	&	\textbf{7.2090}\\
        \toprule
Tong Xiangyu	&	$9.05_{\pm 0.12}^*$	&	$9.20_{\pm 0.11}^*$	&	$\textbf{9.42}_{\pm 0.11}$	&	$15.50_{\pm 0.14}$	&	$15.23_{\pm 0.15}$	&	$\textbf{16.03}_{\pm 0.14}$	&	$2.93_{\pm 0.09}$	&	$3.18_{\pm 0.1}$	&	$\textbf{3.20}_{\pm 0.11}$	&	$15.11_{\pm 0.14}$	&	$14.81_{\pm 0.14}$	&	$\textbf{15.50}_{\pm 0.14}$	\\
Bai Zhantang	&	$9.38_{\pm 0.16}^*$	&	$9.54_{\pm 0.12}$	&	$\textbf{9.87}_{\pm 0.14}$	&	$15.00_{\pm 0.18}^*$	&	$15.09_{\pm 0.17}$	&	$\textbf{17.26}_{\pm 0.17}$	&	$3.98_{\pm 0.15}$	&	$3.75_{\pm 0.14}$	&	$\textbf{4.14}_{\pm 0.14}$	&	$15.13_{\pm 0.18}$	&	$15.23_{\pm 0.17}$	&	$\textbf{15.84}_{\pm 0.18}$	\\
Lv Xiucai	&	$8.05_{\pm 0.14}$	&	$8.65_{\pm 0.14}$	&	$\textbf{9.19}_{\pm 0.15}$	&	$14.29_{\pm 0.16}$	&	$14.51_{\pm 0.17}$	&	$\textbf{14.65}_{\pm 0.16}$	&	$3.59_{\pm 0.13}$	&	$3.56_{\pm 0.14}$	&	$\textbf{3.98}_{\pm 0.14}$	&	$14.43_{\pm 0.16}$	&	$14.30_{\pm 0.17}$	&	$\textbf{14.82}_{\pm 0.16}$	\\
Zhen Huan	&	$9.28_{\pm 0.14}$	&	$9.71_{\pm 0.14}$	&	$\textbf{10.02}_{\pm 0.13}$	&	$21.07_{\pm 0.16}$	&	$20.62_{\pm 0.15}$	&	$\textbf{22.57}_{\pm 0.15}$	&	$6.01_{\pm 0.14}$	&	$6.20_{\pm 0.14}$	&	$\textbf{6.74}_{\pm 0.15}$	&	$19.79_{\pm 0.15}$	&	$19.63_{\pm 0.14}$	&	$\textbf{20.18}_{\pm 0.15}$	\\
Emperor	&	$10.35_{\pm 0.13}$	&	$10.44_{\pm 0.16}$	&	$\textbf{10.78}_{\pm 0.16}$	&	$18.20_{\pm 0.14}$	&	$18.09_{\pm 0.14}$	&	$\textbf{18.75}_{\pm 0.14}$	&	$3.99_{\pm 0.14}$	&	$4.13_{\pm 0.15}$	&	$\textbf{4.40}_{\pm 0.14}$	&	$17.07_{\pm 0.14}$	&	$17.22_{\pm 0.13}$	&	$\textbf{17.44}_{\pm 0.14}$	\\
Real Person	&	$6.98_{\pm 0.24}^*$	&	$9.29_{\pm 0.26}^*$	&	$\textbf{9.96}_{\pm 0.22}$	&	$17.79_{\pm 0.29}^*$	&	$17.85_{\pm 0.28}$	&	$\textbf{18.20}_{\pm 0.28}$	&	$10.98_{\pm 0.26}$	&	$10.95_{\pm 0.27}$	&	$\textbf{11.02}_{\pm 0.26}$	&	$18.87_{\pm 0.31}^*$	&	$19.03_{\pm 0.29}$	&	$\textbf{19.16}_{\pm 0.29}$	\\
        \midrule
        Avg.    & 8.8486	&	9.4724	&	\textbf{9.8709}	&	16.9746	&	16.8978	&	\textbf{17.9098}	&	5.2476	&	5.2943	&	\textbf{5.5811}	&	16.7338	&	16.7010	&	\textbf{17.1562} \\
        \bottomrule
    \end{tabular}
}
\end{table*}

\subsection{Ablation Study}

\begin{figure}[!t]
  \centering
         \includegraphics[width=0.56\columnwidth]{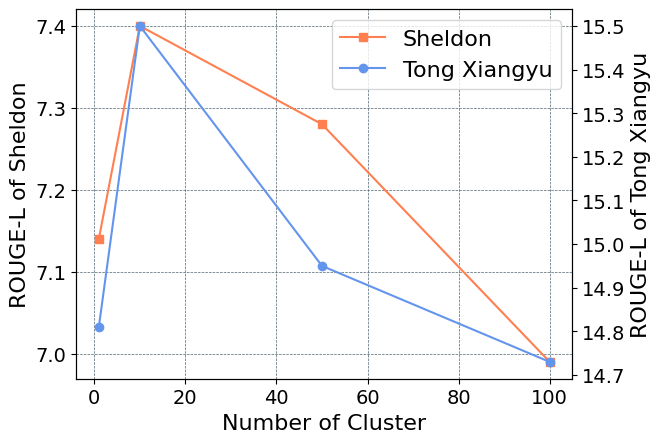}
  \caption{Results of experiments with different numbers of clusters from ROUGE-L (\%).}\label{fig:cluster_num}
\end{figure}

\begin{table}[!t]
    \centering
    \caption{Ablation study results of freezing the general encoder from BLEU and ROUGE-L metrics
(\%).}\label{table:general_encoder}
    \resizebox{0.98\linewidth}{!}{
    \begin{tabular}{lllll}
        \toprule
        \multirow{2}{*}{Responder} & \multicolumn{2}{c}{BLEU} & \multicolumn{2}{c}{ROUGE-L} \\ 
        \cmidrule{2-5}
        ~ & freezing $G_g$ & w/o freezing & freezing $G_g$ & w/o freezing \\
        \midrule
        Sheldon      & ${4.72}_{\pm 0.08}$  & $\textbf{7.16}_{\pm 0.09}$ & $6.85_{\pm 0.09}$ & $\textbf{7.40}_{\pm 0.11}$ \\ 
        \midrule
        Tong Xiangyu & ${6.60}_{\pm 0.11}$ & $\textbf{9.42}_{\pm 0.11}$ & $14.58_{\pm 0.14}$ & $\textbf{15.50}_{\pm 0.14}$ \\ 
        \bottomrule
    \end{tabular}
    }
\end{table}

\textbf{Impact of the querier-contrastive loss and clustering strategy.}
From Table \ref{table:ablation_full_result}, we can observe that introducing querier-contrastive loss results in average improvements of 9.9\% on the English dataset and 2.5\% on the Chinese dataset, demonstrating its role in helping the specific encoder to extract personalized representations for different queriers. Additionally, introducing the query similarity-based clustering strategy yields average improvements of 4.3\% on the English datasets and 2.7\% on the Chinese datasets, revealing its effectiveness in maintaining semantic understanding by focusing contrastive learning within clusters.

\textbf{Impact of the Number of Clusters.}
To explore the impact of the number of clusters, which is a hyperparameter, we conduct an ablation experiment by setting different cluster numbers during data processing and training. The results are shown in Figure \ref{fig:cluster_num}. We observe that when the number of clusters is set to 10 or 50, incorporating the query similarity-based clustering strategy yields better results than not using the clustering strategy. However, as the number of clusters increases further, the training process becomes unstable, and model performance decreases.

\textbf{Impact of the General Encoder.} 
We further conduct experiments by freezing the general encoder, which is initialized with a pretrained LLM, during training to investigate its contribution. The results are presented in Table \ref{table:general_encoder}. Compared with freezing the general encoder, not freezing yields relative improvements of 51.7\% and 8.0\% in BLEU and ROUGE-L scores, respectively, on the Sheldon dataset, and 42.7\% and 6.3\% on the Tong Xiangyu dataset. These results demonstrate that the general encoder’s ability to capture cross-querier personality and semantic information plays an important role in generating high-quality responses.

\subsection{Visualization}\label{sec:visualization}

\begin{figure}[!t]
  \centering
    \begin{subfigure}{0.48\linewidth}
         \centering
         \includegraphics[width=\textwidth]{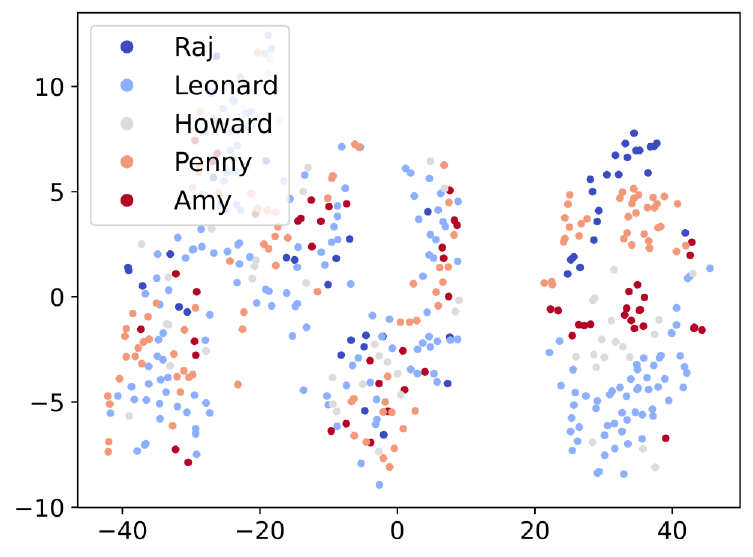}
         \caption{\centering Finetuning}
         \label{fig:personalized_type1}
    \end{subfigure}
    \begin{subfigure}{0.48\linewidth}
         \centering
         \includegraphics[width=\textwidth]{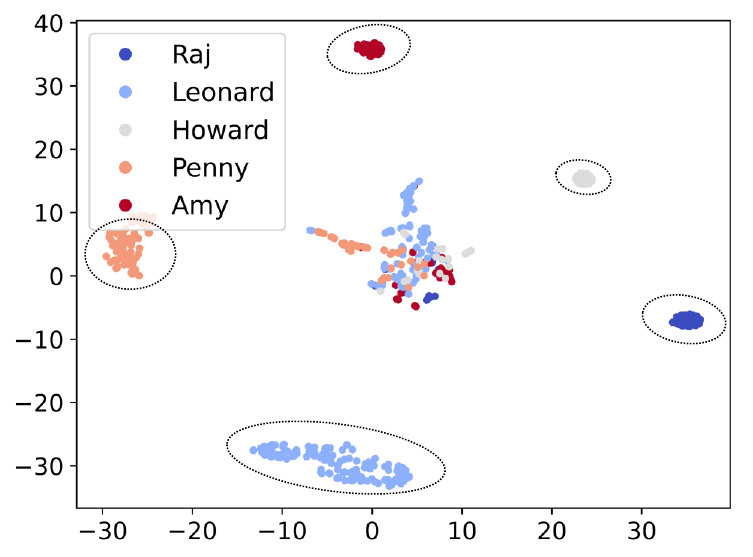}
         \caption{\centering Ours}
         \label{fig:personalized_type2}
    \end{subfigure}
  \caption{English dialogue representations of different queriers interacting with Sheldon.}\label{fig:visualization_bigbang}
  \vspace{-0.8em}
\end{figure}

\begin{figure}[!t]
  \centering
    \begin{subfigure}{0.48\linewidth}
         \centering
         \includegraphics[width=\textwidth]{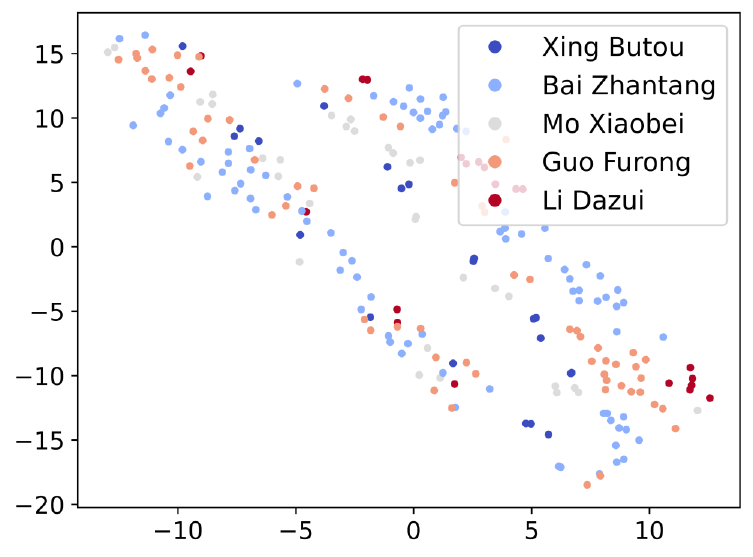}
         \caption{\centering Finetuning}
         \label{fig:visual_ft}
    \end{subfigure}
    \begin{subfigure}{0.48\linewidth}
         \centering
         \includegraphics[width=\textwidth]{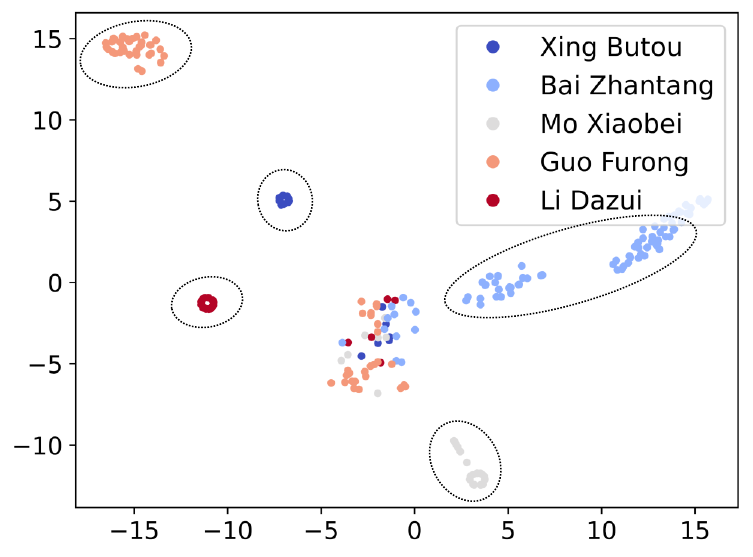}
         \caption{\centering Ours}
         \label{fig:visual_our}
    \end{subfigure}
  \caption{Chinese dialogue representations of different queriers interacting with Tong Xiangyu.}\label{fig:visualization_wulin}
  \vspace{-0.4em}
\end{figure}

\begin{figure*}[!t]
  \centering
    \begin{subfigure}{0.9\linewidth}
         \centering
         \includegraphics[width=\textwidth]{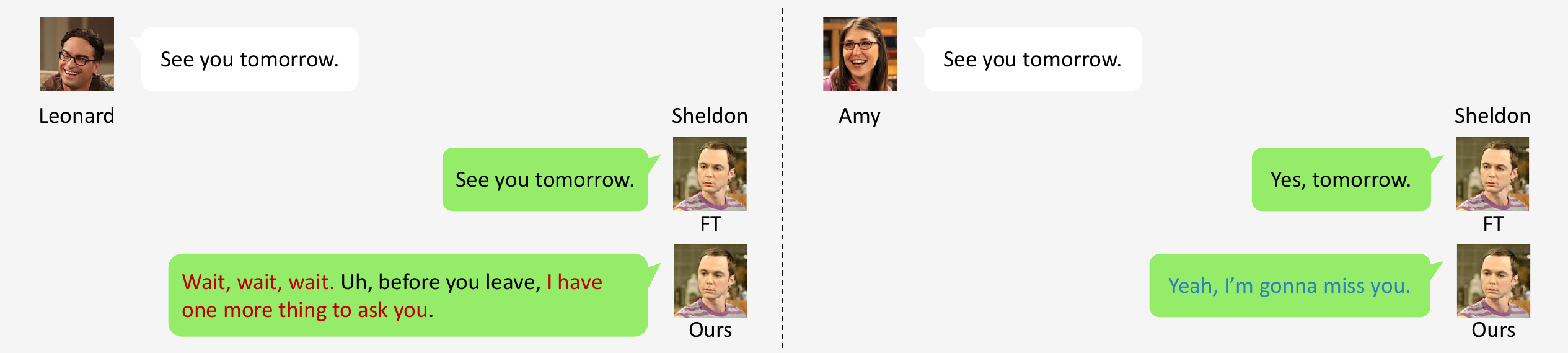}
         \caption{\centering Case 1: Same query asked by Leonard and Amy and answered by Sheldon.}
         \label{fig:case2_en}
    \end{subfigure}
    \\[1ex]
    \begin{subfigure}{0.9\linewidth}
         \centering
         \includegraphics[width=\textwidth]{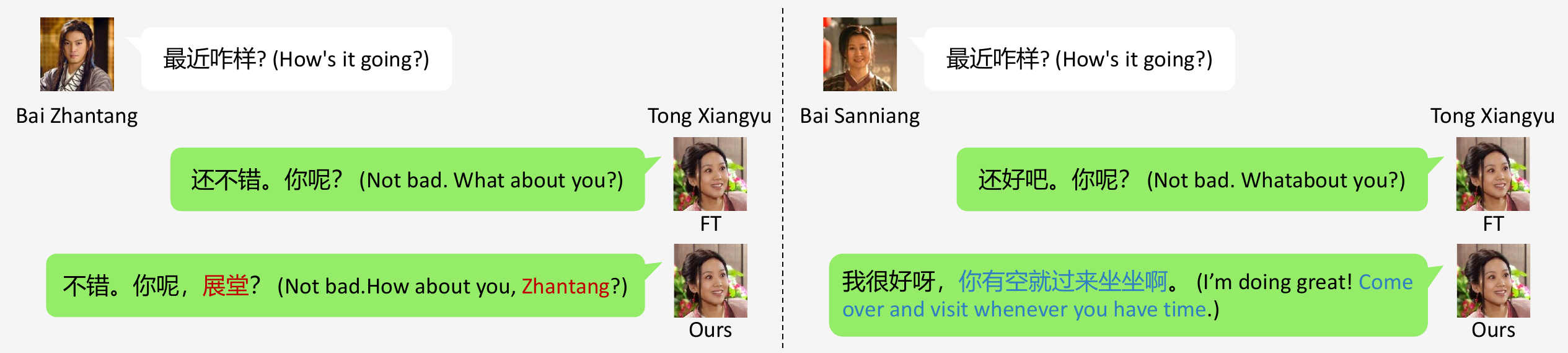}
         \caption{\centering Case 2: Same query asked by Bai Zhantang and Bai Sanniang and answered by Tong Xiangyu}
         \label{fig:case1_zh}
    \end{subfigure}
  \caption{Case study with Sheldon from The Big Bang Theory and Tong Xiangyu from My Own Swordsman.}\label{fig:case1}
  \vspace{-0.4em}
\end{figure*}

To visualize how our design distinguishes between different queriers, we extract the output of the final layer of our specific encoder for samples from one cluster and plot them using t-SNE \cite{tsne}. We introduce FT for comparison. Figure \ref{fig:visualization_bigbang} shows the representation distribution of Sheldon-oriented dialogues from {\em The Big Bang Theory}; and Figure \ref{fig:visualization_wulin} shows the representation distribution of Tong Xiangyu-oriented dialogues dataset from {\em My Own Swordsman}. We can find that, with FT, the representations of dialogues from different queriers are intermixed. In contrast, our design effectively groups the representations of dialogues from the same querier together, distinguishing them from those of different queriers. For our design, a small portion of representations from different queriers clusters together, which may be caused by the dimension reduction of the original high-dimensional features.

\subsection{Case Study}

To take a closer look at the difference between FT and our design, we present several cases in Figure \ref{fig:case1}. We ask the same query with different queriers and compare the responses. In particular, in the case of {\em The Big Bang Theory}, our design demonstrates an understanding of the relationship between the querier and the responder. For Leonard, who is a friend, our design asks him to wait for a moment, while for Amy, who is a romantic partner, our design expresses a sense of missing her. In the case of {\em My Own Swordsman}, our design accurately identifies Bai Zhantang in its response and offers to invite another querier home for a greeting. These cases demonstrate that FT tends to produce similar responses for the same query regardless of the querier. In contrast, our design generates distinct responses by recognizing the querier's identity and the relationship between the querier and the responder. Additional cases can be found in Appendix \ref{sec:case_study}.

\section{Conclusion}
In this work, we have studied the new problem of querier-aware LLM personalization. We have designed a dual-tower model architecture that decouples the responder’s general personality from querier-specific personality. We also have introduced a querier-contrastive loss with multi-view augmentation, along with a query similarity-based dialogue clustering strategy for effectively and efficiently differentiating querier-specific representation. We finally have built the MQDialog dataset and extensively evaluated our design, demonstrating the superiority over several baselines and highlighting the necessity of key design modules. 

\section*{Limitations and Future Work.} 
We construct the dataset and train a querier-aware LLM using dialogues between the responder and various queriers, which serve as the primary data source. However, non-dialogue content from the original scripts, such as scene descriptions and character actions, is filtered out. An interesting direction for future research lies in incorporating additional contextual information, such as the querier’s social relationships, knowledge background, and scene-specific details like time and location, into the LLM’s input. This would serve as a valuable complement to our current dialogue-focused design. 

Furthermore, we focus on generating personalized responses for each individual querier, tailored to their unique personality and relationship with the responder. In our current setting, the number of queriers per responder remains at a moderate scale. Another interesting research direction involves scaling personalization for million-scale groups of queriers. One straightforward extension of our approach to address this challenge is to cluster queriers based on their personal profiles and subsequently train a group-level querier-aware model.

\section*{Ethical Statement.}
This work explores a new form of query-aware LLM personalization. The design, implementation, and use of the Querier-Aware Responder and the proposed MQDialog dataset are guided by ethical principles for responsible AI deployment. (1) Dataset Construction: All data used in this study are either publicly available or anonymized. Additionally, chat records are sourced from an author’s personal chat history and the use of this data has been approved by all participants in the dialogues, only for research purposes. The WeChat dataset is not public. Instead, we provide a toolkit for extracting and cleaning dialogues from messaging apps, such as WhatsApp and WeChat, in Section \ref{sec:data_construct}. (2) Fair Usage: The primary objective of querier-aware personalized LLM is to provide a more realistic and engaging user experience while promoting the harmonious development of AI systems. We believe that the responsible application of our design aligns with ethical guidelines. 

\section*{Acknowledgments}
This work was supported in part by National Key R\&D Program of China (No. 2022ZD0119100), China NSF grant No. 62025204, No. 62202296, No. 62272293, No. 62441236, and No. U24A20326, Alibaba Innovative Research (AIR) Program, SJTU-Huawei Research Program, and Tencent WeChat Research Program. The opinions, findings, conclusions, and recommendations expressed in this paper are those of the authors and do not necessarily reflect the views of the funding agencies or the government.

\section*{GenAI Usage Disclosure}
For automated winning rate evaluation, we employ GPT-4-turbo-2024-04-09 to determine which of two responses, generated by different models, better reflects what the responder would likely say to a specific querier. The total API cost for this evaluation is approximately \$51.8. To assess the alignment between LLM-based judgments and human judgments, we also conducted a human evaluation, with details provided in Section \ref{sec:win}. Additionally, we use GPT-4o for generating querier characteristics, with the corresponding instruction set outlined in Table \ref{table:instruction_for_qpg}. All characteristic prompts produced by GPT-4o are manually reviewed to ensure their accuracy and semantic completeness.

\bibliographystyle{ACM-Reference-Format}
\bibliography{sample-base}

\clearpage
\appendix

\begin{table}[!t]
    \centering
    \caption{Source links of scripts used in MQDialog dataset.}\label{table:source_link}
    \resizebox{\linewidth}{!}{
    \begin{tabular}{l|p{0.8\linewidth}}
    \toprule
        Corpus & Source Link \\ 
        \midrule
        The Big Bang Theory & \href{https://bigbangtrans.wordpress.com/series-1-episode-1-pilot-episode}{https://bigbangtrans.wordpress.com/series-1-episode-1-pilot-episode} \\
        Friends & \href{https://fangj.github.io/friends}{https://fangj.github.io/friends} \\
        Modern Family & \href{https://plexuspictures.com/web/wp-content/uploads/2015/01/Modern-Family-pilot-script-aka-My-American-Family.pdf}{https://plexuspictures.com/web/wp-content/uploads/2015/01/Modern-Family-pilot-script-aka-My-American-Family.pdf} \\
        My Own Swordsman & \href{https://m.juben.pro/writing/4-34-1-ccontent-hp.html}{https://m.juben.pro/writing/4-34-1-ccontent-hp.html} \\
        Empresses in the Palace & \href{https://github.com/KMnO4-zx/huanhuan-chat/tree/master/dataset/input/huanhuan}{https://github.com/KMnO4-zx/huanhuan-chat/tree/master/dataset/input/huanhuan} \\ 
    \bottomrule
    \end{tabular}
    }
\end{table}

\section{Data Source Explanation}\label{sec:data_source_link}

The scripts are collected from the Internet and the source links are listed in Table \ref{table:source_link}. 
As for the WeChat dataset, one of the authors serves as the responder and the queriers are colleagues or teachers from the same lab, or a colleague from a partner organization. They also agree that the dataset can be properly used for research purposes. 
A toolkit for extracting and cleaning dialogues from messaging apps, such as WhatsApp and WeChat, is introduced in Section \ref{sec:data_construct}. As long as both parties in the dialogue agree to use the data, we can construct the dataset.  

\section{Prompt for Evaluation}\label{sec:eval_prompt}
Figure \ref{table:eval_prompt_en} and Figure \ref{table:eval_prompt_zh} show detailed prompts used for evaluating the winning rate. First, we prompt the LLM to determine which of the two responses better aligns with the responder's personality and resembles how the responder would naturally speak to the specific querier. We request the LLM to provide its reasoning process. Next, we prompt the LLM to format its output based on the reasoning process.

\section{Case Study}\label{sec:case_study}

\begin{figure*}[!t]
  \centering
    \begin{subfigure}{0.87\linewidth}
         \centering
         \includegraphics[width=\textwidth]{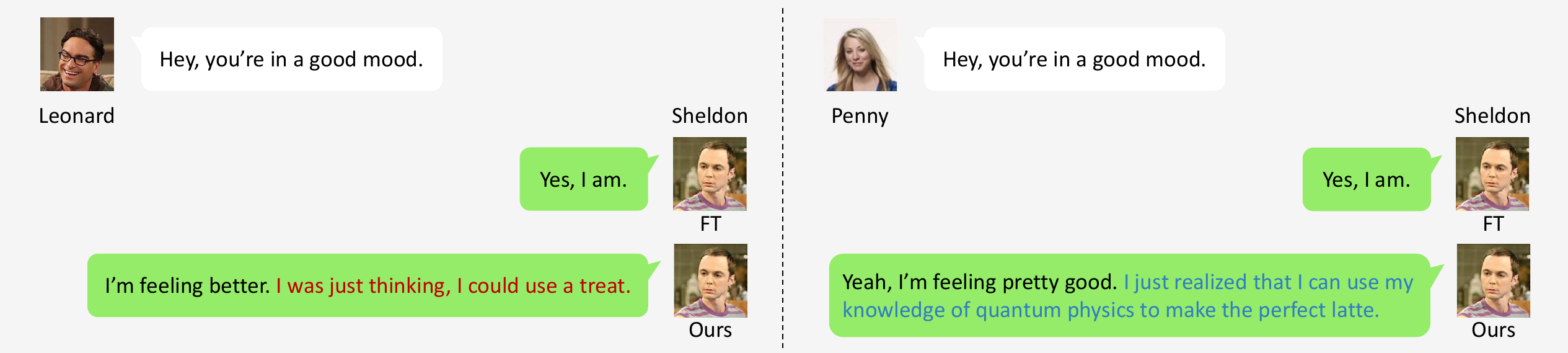}
         \caption{\centering More Case 3: Same query asked by Leonard and Penny and answered by Sheldon.}
         \label{fig:case4_en}
    \end{subfigure}
    \\[1ex]
    \begin{subfigure}{0.87\linewidth}
         \centering
         \includegraphics[width=\textwidth]{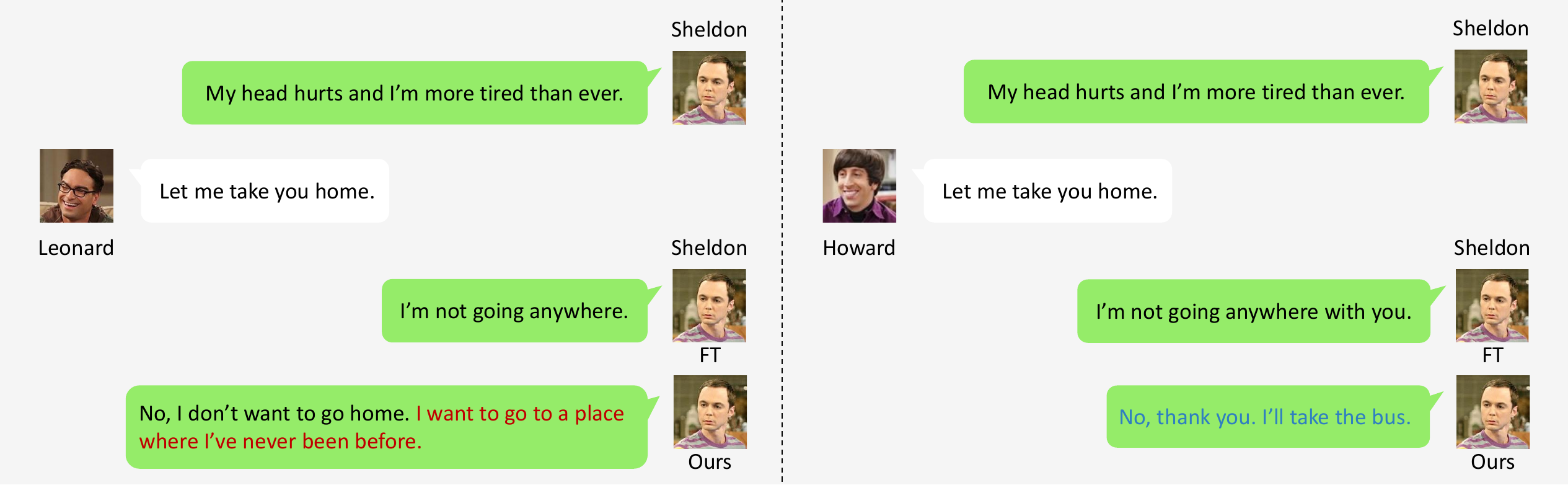}
         \caption{\centering More Case 4: Same query asked by Leonard and Howard and answered by Sheldon.}
         \label{fig:case6_en}
    \end{subfigure}
    \vspace{-0.5em}
  \caption{Case study with Sheldon from The Big Bang Theory.}\label{fig:case}
\end{figure*}

\begin{figure*}[!t]
  \centering
    \begin{subfigure}{0.87\linewidth}
         \centering
         \includegraphics[width=\textwidth]{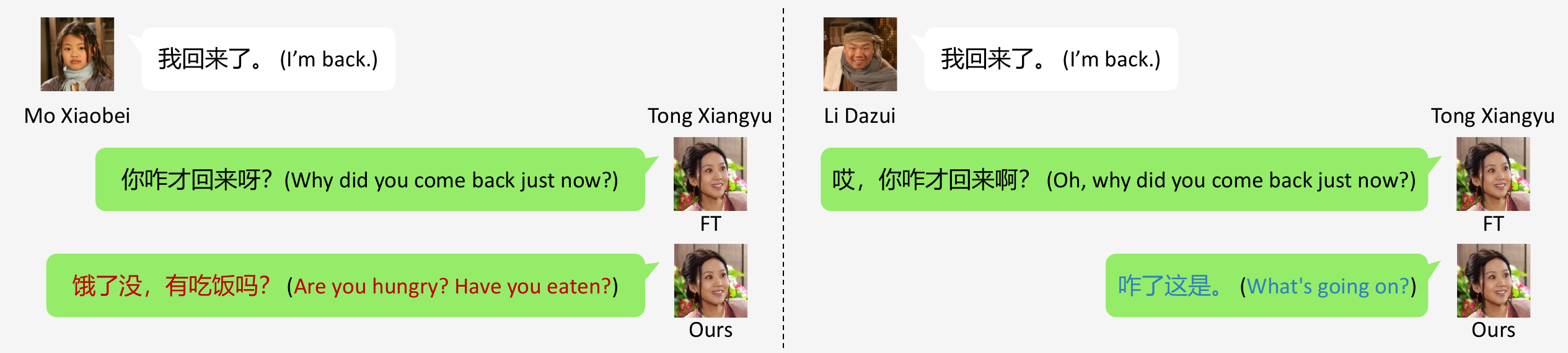}
         \caption{\centering More Case 5: Same query asked by Mo Xiaobei and Li Dazui and answered by Tong Xiangyu.}
         \label{fig:case3_zh}
    \end{subfigure}
    \\[1ex]
    \begin{subfigure}{0.87\linewidth}
         \centering
         \includegraphics[width=\textwidth]{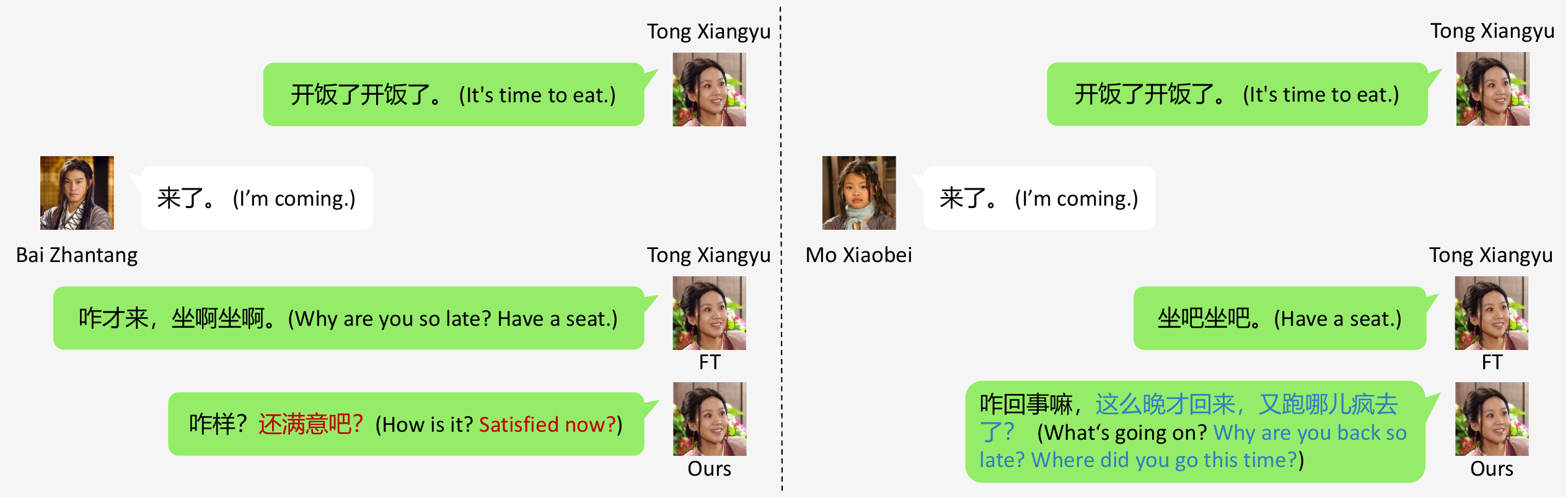}
         \caption{\centering More Case 6: Same query asked by Bai Zhantang and Mo Xiaobei and answered by Tong Xiangyu.}
         \label{fig:case5_zh}
    \end{subfigure}
    \vspace{-0.5em}
  \caption{Case study with Tong Xiangyu from My Own Swordsman.}\label{fig:more_case}
  \vspace{-0.5em}
\end{figure*}

We present several cases in Figure \ref{fig:case} and Figure \ref{fig:more_case}, including examples of multi-round interactions. We ask the same query with different queriers and compare the responses. We can observe that FT often produces similar responses to the same query, regardless of who asks. In contrast, our approach generates tailored responses by considering the querier’s identity and their relationship with the responder.

\section{Discussion of Our Design}

\subsection{Evidence for low-rank design}
As shown in Figure \ref{fig:model}, we adopt low-rank matrices for the feedforward layer in the specific block and we can explain this design in terms of design rationale and empirical evaluation. 
From design rationale, in conventional adaptation scenarios of LLM, a large pretrained model is typically trained for a downstream task by adding a small number of parameters, such as using an adapter \cite{adapterfusion}. In our design, the responder's general personality is dominant, so we use a full-parameter encoder to capture it. For encoding the influence of the querier's personality on the response, we employ a low-rank encoder. Additionally, different from the mainstream paradigm of coupling the adapter and the general block, our design decouples the specific encoder from the general encoder using a dual-town structure.  
From empirical evaluation, the improvement over the baselines has validated the effectiveness of this design. The visualization results in Section \ref{sec:visualization} also reveal that the dialogue representations for different queriers can be clearly distinguished using the low-rank specific encoder.

\subsection{Cold-start for new queriers}
For a completely new querier with no prior interaction history, their current query will be fed into both the general and specific encoders of the proposed dual-tower model to generate a response. Since these two encoders are shared across all queriers, our design allows the query understanding capabilities learned from interactions with previous queriers to benefit this new user. However, to obtain personalized and better responses for the new querier, we need to collect the conversations between the new querier and the responder, and use this data to train the model.

\begin{figure*}[!t]
  \centering
    \begin{tcolorbox}[title=Step 1: Judge which answer better meets our requirements.]
\footnotesize
\textbf{\textit{System}}: \\
You need act as the \{RESPONDER\} in the \{DATASET SOURCE\} and determine which of the two responses to \{QUERIER\} is better.   \\
\textbf{\textit{Message}}: \\
As a \{RESPONDER\}, please judge which of the following two responses is more authentic, aligns better with your personality, resembles what you would say to \{QUERIER\}, and fits your relationship with \{QUERIER\}. \\
The better response should be a semantically complete sentence and should especially not include any unrelated information. \\
Here are some examples from your previous conversations with \{QUERIER\}: \\
~ \\
\{FEW-SHOT EXAMPLES\}  \\
~ \\
The current dialogue between you and \{QUERIER\} is as follows:  \\
Current dialogue: \{DIALOGUE\} \\
A: \{RESULT 1\} \\
B: \{RESULT 2\} \\
\end{tcolorbox}

\begin{tcolorbox}[title=Step 2: Format the output to indicate which answer wins based on the reasoning process.]
\footnotesize 
\textbf{\textit{System}}: \\ 
You are a teacher who grades multiple-choice queries.    \\
\textbf{\textit{Message}}:  \\
Please determine whether the answer derived from the reasoning is "A" or "B" based on the reasoning information. \\
The following is the reasoning process and result: \\
~ \\
 \{REASONING\} \\
 ~ \\
Your answer should be only "A" or "B". \\
\end{tcolorbox}

    \caption{Prompt template for evaluating the winning rate for the English datasets.}\label{table:eval_prompt_en}
\end{figure*}

\begin{figure*}[!t]
  \centering
    \begin{tcolorbox}[title=Step 1: \chinese{判断哪个回答更符合我们的要求}]
\footnotesize
    \textbf{\textit{System}}: \\
    \chinese{你将扮演\{DATASET SOURCE\}中的\{RESPONDER\}，判断你对\{QUERIER\}的两个回复哪个更好。}    \\
    \textbf{\textit{Message}}: \\
    \chinese{作为\{RESPONDER\}，请判断以下两个回答中，哪个更真实，更贴近你的性格，更像是你会对\{QUERIER\}说的话，更符合你跟\{QUERIER\}之间的关系。}\\
    \chinese{更好的回答是语义完整的一句话，特别是不应有其他不相关的信息出现。}\\
    \chinese{参考你跟\{QUERIER\}之前对话的一些例子：} \\ 
    ~ \\
    \{FEW-SHOT EXAMPLES\}  \\
    ~ \\
    \chinese{以下是你与\{QUERIER\}当前对话的两个回答} \\
    \chinese{当前对话：\{DIALOGUE\}} \\
    A: \{RESULT 1\} \\
    B: \{RESULT 2\} \\
\end{tcolorbox}

\begin{tcolorbox}[title=Step 2: \chinese{根据推理过程，格式化输出结果}]
\footnotesize 
\textbf{\textit{System}}: \\ 
\chinese{你是一位选择题的批改老师。}   \\
\textbf{\textit{Message}}: \\ 
\chinese{请根据推理信息，判断出推理所得到的答案是"A"还是"B"。} \\
\chinese{以下为推理过程与结果：} \\
~\\ 
\{REASONING\} \\
~\\
\chinese{你的回答应仅为"A"或"B"。} \\
\end{tcolorbox}

\caption{Prompt template for evaluating the winning rate for the Chinese datasets.}\label{table:eval_prompt_zh}
\end{figure*}

\section{Training Cost and Efficiency}
As introduced in Section \ref{sec:approach}, the general block in the dual-tower model is initialized from a pretrained LLM. The specific block consists of two low-rank matrices and a multi-head attention layer, which shares parameters with the multi-head attention layer of the general block. Thus, the additional parameters are only the low-rank matrices, accounting for approximately 1\% of pretrained LLM size. Specifically, for 7B LLM on Nvidia V100, finetuning for 200 steps takes 1.19 GPU hours, while training for 200 steps with our method takes 1.36 GPU hours. Therefore, our method is computationally efficient and comparable to standard finetuning in terms of training cost.

\end{document}